\documentclass[11pt]{article}

\usepackage[margin=1in]{geometry}
\usepackage{microtype}
\usepackage{amsmath,amssymb,mathtools}
\usepackage{booktabs}
\usepackage{array}
\usepackage{multirow}
\usepackage{graphicx}
\usepackage{xcolor}
\usepackage{tcolorbox}
\tcbuselibrary{breakable}
\usepackage{tikz}
\usetikzlibrary{arrows.meta,positioning,fit}
\usepackage{enumitem}
\usepackage{hyperref}
\usepackage{url}
\usepackage{caption}
\usepackage{subcaption}
\usepackage{float}

\usepackage{longtable}
\usepackage{tabularx}
\usepackage{siunitx}
\usepackage{authblk}
\usepackage[section]{placeins}
\hypersetup{
  pdftitle={QureRadEmbed: Structuring Radiological Similarity through Attribute and Reasoning Supervision},
  pdfauthor={Janhavi Prabhu, Sahil, Shivam Ashok Shukla, Manoj Tadepalli},
  colorlinks=true,
  linkcolor=blue!55!black,
  citecolor=blue!55!black,
  urlcolor=blue!55!black
}

\definecolor{boxblue}{RGB}{238,246,255}
\definecolor{boxblueframe}{RGB}{75,120,170}
\definecolor{boxgreen}{RGB}{239,249,242}
\definecolor{boxgreenframe}{RGB}{70,135,90}
\definecolor{boxorange}{RGB}{255,247,235}
\definecolor{boxorangeframe}{RGB}{180,115,45}
\definecolor{boxgray}{RGB}{247,247,247}
\definecolor{boxgrayframe}{RGB}{130,130,130}

\newtcolorbox{intuitionbox}[1][]{
  colback=boxblue,
  colframe=boxblueframe,
  title=Intuition,
  fonttitle=\bfseries,
  breakable,
  #1
}

\newtcolorbox{retrievalbox}[1][]{
  colback=boxgreen,
  colframe=boxgreenframe,
  title=Sample retrieval,
  fonttitle=\bfseries,
  #1
}

\newtcolorbox{designintuitionbox}[1][]{
  colback=boxorange,
  colframe=boxorangeframe,
  title=Design intuition,
  fonttitle=\bfseries,
  breakable,
  #1
}

\newtcolorbox{caveatbox}[1][]{
  colback=boxgray,
  colframe=boxgrayframe,
  title=Evaluation caveat,
  fonttitle=\bfseries,
  breakable,
  #1
}

\newcommand{\cosim}{\operatorname{cos}}

\newcommand{\etl}{\mathrm{ETL}}

\title{QureRadEmbed: Structuring Radiological Similarity\\
through Attribute and Reasoning Supervision}

\author{Janhavi Prabhu\textsuperscript{*}\quad Sahil\textsuperscript{*}\\[0.25em]
Shivam Ashok Shukla\textsuperscript{$\dagger$}\quad Manoj Tadepalli\textsuperscript{$\ddagger$}}
\affil{Research \& Development, Qure.ai}
\date{}

\begin{document}
\maketitle

\begingroup
\let\footnotesize\small
\renewcommand{\thefootnote}{}
\footnotetext{\textsuperscript{*}Janhavi Prabhu and Sahil are co-first authors.
\textsuperscript{$\dagger$}Shivam Ashok Shukla is the second author.
\textsuperscript{$\ddagger$}Manoj Tadepalli is the corresponding author.}
\endgroup

\begin{abstract}
Radiological similarity depends on disease relationships and on fine details such as laterality, lobe, severity, size, and certainty. Broad biomedical similarity can overlook these qualifiers, particularly when several attributes vary together. We introduce QureRadEmbed, a 4B radiology-aware encoder trained with two complementary signals: RadSim supplies deterministic, attribute-decomposed ranking targets, while RadThought aligns reports with hierarchical evidence and reasoning descriptions. A three-stage curriculum combines these signals with report triplets, finding perturbations, and single- and cross-attribute contrasts. The final model achieves $0.996$ mean ordering accuracy across ten controlled synthetic attributes and raises Spearman correlation with the designed joint-attribute targets from $0.501$ to $0.976$. On external findings-to-impression retrieval, Recall@1 reaches $10.5\%$ on Open-I, $12.4\%$ on testing XR, and $42.9\%$ on testing CT, compared with $6.6\%$, $5.7\%$, and $31.4\%$ for its backbone. Frozen embeddings support finding extraction with only 100 labeled testing-XR reports (macro-F1 $0.481$ versus $0.412$ for the backbone). Whole-report comparison costs 8.8 seconds per 1,000 pairs in our benchmark, versus 2,755.1 seconds for the generative evaluator GREEN. Sentence-level comparison improves sensitivity to local discrepancies, although generative evaluation remains stronger on several expert-rated and subtle-error tasks. The results support reusable radiology-aware representations for search, structured report indexing, and efficient report comparison.
\end{abstract}

\section{Introduction}
A useful radiology embedding must recognize both \emph{what is related} and \emph{what is clinically different}. Emphysema and COPD share a disease family despite different wording. Conversely, ``right upper lobe nodule'' and ``left lower lobe nodule'' share most words but refer to different anatomy. Severity and size introduce graded distinctions: mild versus moderate disease should not be treated like mild versus severe disease. These properties matter for similar-case search, retrieval-augmented generation, cohort identification, structured report indexing, and comparison of generated reports with reference text.

Biomedical pretraining supplies vocabulary and context, while entity alignment and general retrieval objectives emphasize synonymy or broad relevance \cite{lee2020biobert,gu2021pubmedbert,liu2021sapbert,jin2023medcpt}. Neither objective specifies how strongly a side, location, or severity mismatch should affect similarity. Our initial probes exposed this gap: the Qwen3-Embedding-4B backbone achieved $0.521$ size-triplet accuracy and $0.661$ laterality accuracy. High report similarity could therefore coexist with disagreement on the very attributes needed to select a clinically appropriate match.

Large language models can perform detailed report comparison. GREEN identifies and explains clinically significant errors rather than relying on lexical overlap \cite{ostmeier2024green}. Its flexibility has an inference cost: in our evaluation, GREEN required 2,755.1 seconds per 1,000 report pairs, compared with 8.8 seconds for QureRadEmbed whole-report cosine. This approximately $313\times$ difference motivates an encoder whose report vectors can be cached and reused across many comparisons. It does not imply equivalent judgment: GREEN achieves higher within-study agreement with radiologists on ReXVal ($\tau_b=0.445$ versus $0.336$ for QureRadEmbed cosine). The practical aim is an efficient representation that captures useful clinical distinctions, with richer comparison available when needed.

We address this aim through \textbf{RadSim}, a scoring policy combining attribute agreement, disease-family relationships, and explicit contradiction rules. Intern-S2 extracts stated attributes; deterministic code assigns graded targets. \textbf{RadThought} supplies complementary hierarchical supervision, linking findings to logical evidence and progressively composed reasoning descriptions. It teaches alignment across longer, reasoning-rich text rather than merely increasing the input limit. Single-attribute contrasts target individual distinctions, while cross-attribute examples specify priorities intended to reduce masking of a consequential mismatch by agreement on another attribute.

\begin{designintuitionbox}
Recognizing ``the same disease'' is only the starting point. A useful match must preserve where it is, on which side, how severe or large it is, and how confidently it is described. Related findings should remain close without making their attributes interchangeable.
\end{designintuitionbox}

Our methodological contribution is a framework for constructing and composing radiological similarity supervision. First, RadSim converts extracted clinical attributes into inspectable ranking targets that express graded agreement, contradiction, and disease relatedness. Second, RadThought couples report observations with evidence and reasoning descriptions at multiple levels of abstraction. Third, a curriculum progresses from report alignment to isolated attribute contrasts and then joint-attribute priorities, addressing interaction failures that remain hidden by strong single-attribute accuracy. These supervision streams train a shared encoder through complementary pairwise and triplet-ranking objectives. External retrieval, frozen-encoder probes, and report-comparison experiments evaluate the resulting representation, including the trade-offs between attribute sensitivity and broader semantic similarity.

\FloatBarrier
\subsection{Related Work}
\subsubsection{General-purpose sentence and retrieval embeddings}
Sentence-BERT made semantic comparison efficient by encoding texts independently and comparing their vectors, enabling reuse of document representations across queries \cite{reimers2019}. SimCSE showed that contrastive training with dropout-based views, or natural-language-inference supervision, produces effective sentence representations \cite{gao2021simcse}. E5 extends contrastive representation learning through large-scale weakly supervised text pairs, supporting retrieval, clustering, and classification with a common embedding \cite{wang2022e5}. BGE-M3 broadens this approach across languages, input lengths, and retrieval representations, jointly supporting dense, sparse, and multi-vector scoring through self-knowledge distillation \cite{chen2024bgem3}. These developments establish strong general-purpose encoders, but their broad relevance objectives do not directly specify the relative clinical importance of a side, severity, or measurement mismatch.

The training objective also affects which distinctions are expressed in cosine space. CoSENT optimizes consistency between target-score ordering and embedding similarity \cite{huang2024cosent}; AnglE addresses optimization difficulties associated with cosine saturation \cite{li2023angle}. Our use of CoSENT follows the ranking perspective: RadSim supplies an interpretable ordering of clinical agreement rather than prescribing an absolute cosine for every report pair. Attribute-specific triplets complement that global ordering with controlled local comparisons.

\subsubsection{Instruction tuning and LLM-derived embedding supervision}
INSTRUCTOR conditions embeddings on instructions describing the task and domain, allowing one encoder to express different notions of similarity \cite{su2023instructor}. E5-Mistral demonstrates how LLM-generated examples and contrastive fine-tuning can adapt a decoder backbone for embedding tasks \cite{wang2024llmembeddings}. Gecko similarly distills LLM knowledge into a retriever through synthetic query--passage pairs and LLM relabeling of retrieved positives and hard negatives \cite{lee2024gecko}. Qwen3-Embedding combines large-scale pretraining, supervised adaptation, synthetic data, and model merging to learn multilingual embeddings and rerankers \cite{qwen3embedding}.

QureRadEmbed builds on Qwen3-Embedding and shares the use of generated supervision, but gives the teacher a more constrained role. InternS2 extracts stated clinical attributes; a deterministic policy assigns the numerical target. This separates language understanding from decisions about contradiction, partial agreement, and disease-family relatedness. RadThought supplies a different form of supervision: agreement across report-derived evidence and successive levels of synthesis. Its purpose is alignment of reasoning-rich descriptions, rather than instruction-conditioned retrieval. Our current encoder uses no task prefix; task-conditioned attribute weighting remains a complementary direction.

\subsubsection{Bidirectional encoders, long context, and pplx-embed}
LLM2Vec converts decoder-only models into text encoders by enabling bidirectional attention, training masked next-token prediction, and applying unsupervised contrastive learning \cite{behnamghader2024llm2vec}. \textbf{pplx-embed} takes a related architectural direction through diffusion-based pretraining followed by multi-stage contrastive learning \cite{eslami2026pplx}. Its bidirectional backbone supports mean pooling; the family includes both standard dense embeddings and contextual embeddings that retain document-level information within passage representations. This makes pplx-embed especially relevant to radiology, where a local finding can depend on surrounding anatomy, comparisons, and explanatory context. We cite it as related work, not as an evaluated baseline in the present experiments.

Long input capacity and preservation of local detail are distinct problems. Late chunking first contextualizes tokens across a long document and then pools within chunk boundaries, preserving surrounding information that independent chunk encoding can discard \cite{gunther2024latechunking}. ColBERT instead retains token-level vectors and performs late interaction between separately encoded queries and documents \cite{khattab2020colbert}. These methods provide alternatives to compressing every detail into one vector. Our sentence-alignment analyses examine the same practical concern at sentence granularity, while RadThought addresses the supervision of evidence across levels of abstraction. Combining clinical attribute supervision with contextual chunks or richer local interaction is a natural extension, not a capability established by the current pooled encoder alone.

\subsubsection{Biomedical concept and clinical text representations}
BioBERT and PubMedBERT show the value of biomedical language-model pretraining for domain vocabulary and contextual representation \cite{lee2020biobert,gu2021pubmedbert}. SapBERT explicitly aligns synonymous biomedical entities using terminology structure \cite{liu2021sapbert}. BioLORD grounds concept representations in definitions and ontology-derived descriptions; BioLORD-2023 further combines LLM-generated descriptions with clinical knowledge-graph information \cite{remy2024biolord}. MedCPT learns biomedical retrieval from large-scale PubMed search logs, supplying relevance supervision beyond language modeling or synonym alignment \cite{jin2023medcpt}.

These approaches cover complementary aspects of medical meaning: terminology, concept relatedness, sentence semantics, and information retrieval. Radiology adds a requirement to preserve graded differences \emph{within} a finding. Two descriptions can identify the same concept yet differ in lobe, side, count, severity, or certainty. RadSim connects disease-family relatedness with those explicit attributes, while controlled contrasts and joint-axis targets determine how attribute changes influence similarity. The distinction concerns the supervision provided, rather than an assumption that existing biomedical encoders cannot represent such details.

\subsubsection{Radiology-specific representation learning}
RadBERT adapts transformer language models to radiology reports and evaluates their utility for applications including abnormal-sentence classification, report coding, and summarization \cite{yan2022radbert}. BioViL emphasizes radiology-specific text semantics through CXR-BERT within a biomedical vision--language framework \cite{boecking2022biovil}. RadEval introduces a radiology text-evaluation framework and RadEvalModernBERT as a domain-specific representation model \cite{xuradeval}. Together, these studies support specialization to the language and structure of radiology rather than reliance on general biomedical vocabulary alone.

Factually aware contrastive training and radiology-specific retrieval provide further methodological precedents. RadBERT-CL constructs contrastive examples that preserve or alter factual clinical information, while RadSearch learns report retrieval from findings--impression supervision \cite{jaiswal2021radbertcl,savage2025radsearch}. QureRadEmbed develops this supervision-based approach through graded attribute targets, alignment across report-derived reasoning levels, and controlled examples specifying how attributes interact.

Our emphasis is an explicit similarity policy spanning disease relationships, graded attributes, and their interactions. Location-controlled examples separate laterality from zone differences, and cross-axis pairs address the possibility that a magnitude match can obscure a spatial contradiction. Frozen finding/status probes and attribute-specific linear heads then test whether the learned representation exposes clinically useful information beyond its aggregate retrieval score. These analyses complement radiology-domain pretraining and image--text alignment rather than replacing them.

\subsubsection{Clinical report comparison and evaluation}
CheXbert represents report-level finding status, while RadGraph extracts clinical entities and relations \cite{smit2020chexbert,jain2021radgraph}. RaTEScore uses entity-aware comparisons to accommodate medical synonyms and negation \cite{ratescore}. GREEN uses generative models to identify and explain clinically significant report errors, and RadSEM performs finding-level clinical consistency assessment \cite{ostmeier2024green,radsem}. These approaches can preserve distinctions that a single report cosine compresses, motivating our comparison of whole-report embeddings, local alignment, and directional coverage.

MTEB evaluates embeddings across multiple task families, emphasizing that retrieval, classification, and semantic similarity are related but different capabilities \cite{muennighoff2023mteb}. We adopt this broader evaluation principle in a radiological setting: external retrieval tests clinical matching, controlled interventions test attribute sensitivity, linear probes test accessible finding information, and expert-rated comparisons test report-quality agreement. QureRadEmbed targets an efficient reusable representation across these uses; it does not assume that stronger retrieval or attribute ordering entails superior agreement with every clinical quality judgment.

\FloatBarrier
\subsection{Problem Formulation}
\label{sec:problem}
For radiological text $x$, the encoder produces a unit vector $f_\theta(x)\in\mathbb{R}^{2560}$ and similarity $s_\theta(x,y)=f_\theta(x)^\top f_\theta(y)$. The desired structure has three levels: disease-family relatedness, agreement on individual attributes, and the degree of disagreement along graded attributes. ``No pneumothorax'' should not be a good match for ``pneumothorax'' simply because the diagnostic word is shared; equally, nearby lesion sizes should remain more similar than substantially different sizes.

An attribute intervention can be pictured as a displacement
\[
\Delta_a(x)=f_\theta(T_a x)-f_\theta(x),
\]
where $T_a$ changes only attribute $a$. We want each attribute to have a distinguishable effect that remains accessible when other attributes change. This is a training objective, not an assumption that attributes form fixed, mutually orthogonal directions: their effects can depend on the finding and surrounding text. A single cosine score blends these contributions. A large response to size or lobe location can therefore obscure a side mismatch even when isolated laterality tests are accurate. Controlled single-axis data hold other canonical attributes fixed; joint-axis data specify their intended relative influence.

Report templates further compress the useful similarity range. A positive at $0.89$ and a near-miss at $0.88$ are both highly similar, but their order determines retrieval. We therefore learn relative rankings and use task-specific margins rather than require a universal similarity cutoff. Retrieval is one application of this geometry; attribute probes and report-comparison tasks test whether its distinctions are clinically useful.

\section{Methodology}
\label{sec:methods}
\subsection{Data Preparation and Supervision}
\label{sec:data}
\subsubsection{Report collection and training examples}
The training corpus contains approximately $1.1$M chest-radiograph reports and $174$k CT/CTA reports retained after parsing and abnormality filtering. Each report is organized into findings, impression, abnormalities, and prominence. Findings provide detailed observations, while impressions summarize the clinically salient conclusions; retaining both supports supervision at sentence, section, and whole-report levels. Structured abnormality labels support the selection of clinically similar reports and hard negatives.

The training examples serve complementary purposes. Sentence pairs emphasize local clinical distinctions; findings and impression pairs teach graded similarity over longer descriptions. Report and sentence triplets teach the encoder to rank a compatible description above a clinically different one. Controlled attribute contrasts isolate specific changes, while perturbation and reordering examples distinguish changes in clinical content from changes in presentation. RadThought adds hierarchical reasoning and longer contexts.

Supervision has two forms: graded pairs for CoSENT and anchor--positive--negative triplets for cosine ranking. Major families include $4.28$M sentence pairs, $6.34$M findings pairs, $6.32$M impression pairs, $4.23$M report triplets, and $1.37$M sentence triplets. Initial size/laterality sets contain approximately $149$k/$52$k triplets; RadThought triplet families contain $135$k--$148$k examples each. Perturbation/reordering adds $1.88$M pairs. Dataset caps and sampling balance these sources; raw corpus size is not their training frequency.

\subsubsection{Attribute extraction and RadSim}
Intern-S2-Preview extracts presence, size, laterality, severity, certainty, spatial correspondence, and morphology \cite{interns2}. The prompt preserves unstated information as missing rather than inferring facts. RadSim then combines deterministic per-finding attribute agreements with disease-family and organ-system adjustments:

\begin{equation}
S_{\text{RadSim}}(x,y)
=\operatorname{clip}_{[-1,1]}\!\left[
\frac{\displaystyle\sum_{f}\sum_{a\in\mathcal{A}_f}w_a\,r_{f,a}(x,y)}
     {\displaystyle\sum_{f}\sum_{a\in\mathcal{A}_f}w_a}
+ B_{\mathrm{family}}(x,y)
+ B_{\mathrm{organ}}(x,y)
+ \epsilon(x,y)
\right].
\label{eq:radsim}
\end{equation}

Here $x$ and $y$ are the two descriptions, $f$ indexes a clinical finding, $a$ indexes an attribute, and $r_{f,a}$ is its agreement score. The set $\mathcal{A}_f$ contains applicable attributes, and $w_a$ controls their contribution to the weighted agreement. Presence and size receive weight $2.5$, severity $1.2$, laterality and spatial correspondence $1.0$, morphology $0.8$, and certainty $0.6$. Explicit absence is distinct from omission: a present--absent contradiction contributes negatively, whereas an unstated attribute is not treated as a confirmed absence. Missing size and morphology are excluded when unstated in both descriptions.

$B_{\mathrm{family}}(x,y)$ is a rule-based disease-family adjustment: it preserves relatedness between distinct but clinically associated findings, such as emphysema and COPD or fibrosis and interstitial lung disease. $B_{\mathrm{organ}}(x,y)$ is a coarser organ-system adjustment, including a $+0.20$ bonus for eligible different findings in the same organ system. These terms prevent a clinically related pair from being treated like two unrelated conditions; they are not learned parameters or additional attribute weights. For the same finding and organ in the schematic example, both adjustments are zero, so the attribute differences determine the score. Finally, $\epsilon(x,y)$ is a small deterministic score perturbation, described below.

For positive sizes $a,b>0$ normalized to common units,
\begin{equation}
s_{\mathrm{size}}(a,b)=1-2\frac{|a-b|}{\max(a,b)}.
\label{eq:radsimsize}
\end{equation}
Thus the score reflects relative size difference rather than shared numerals. A nodule--mass continuum preserves relatedness across the terminology boundary. Content-hash-seeded Gaussian jitter ($\sigma=0.02$, clipped at $\pm3\sigma$) perturbs scores reproducibly and can break ties or reorder nearby scores; the complete score is clipped to $[-1,1]$. These are ranking targets, not calibrated cosine values or probabilities of clinical correctness.

\begin{designintuitionbox}
Separate ``what is stated?'' from ``how should a difference affect similarity?'' Structured extraction preserves the evidence, while deterministic scoring makes the supervision policy inspectable and adjustable.
\end{designintuitionbox}

\begin{figure}[!htbp]
\centering
\includegraphics[width=\linewidth]{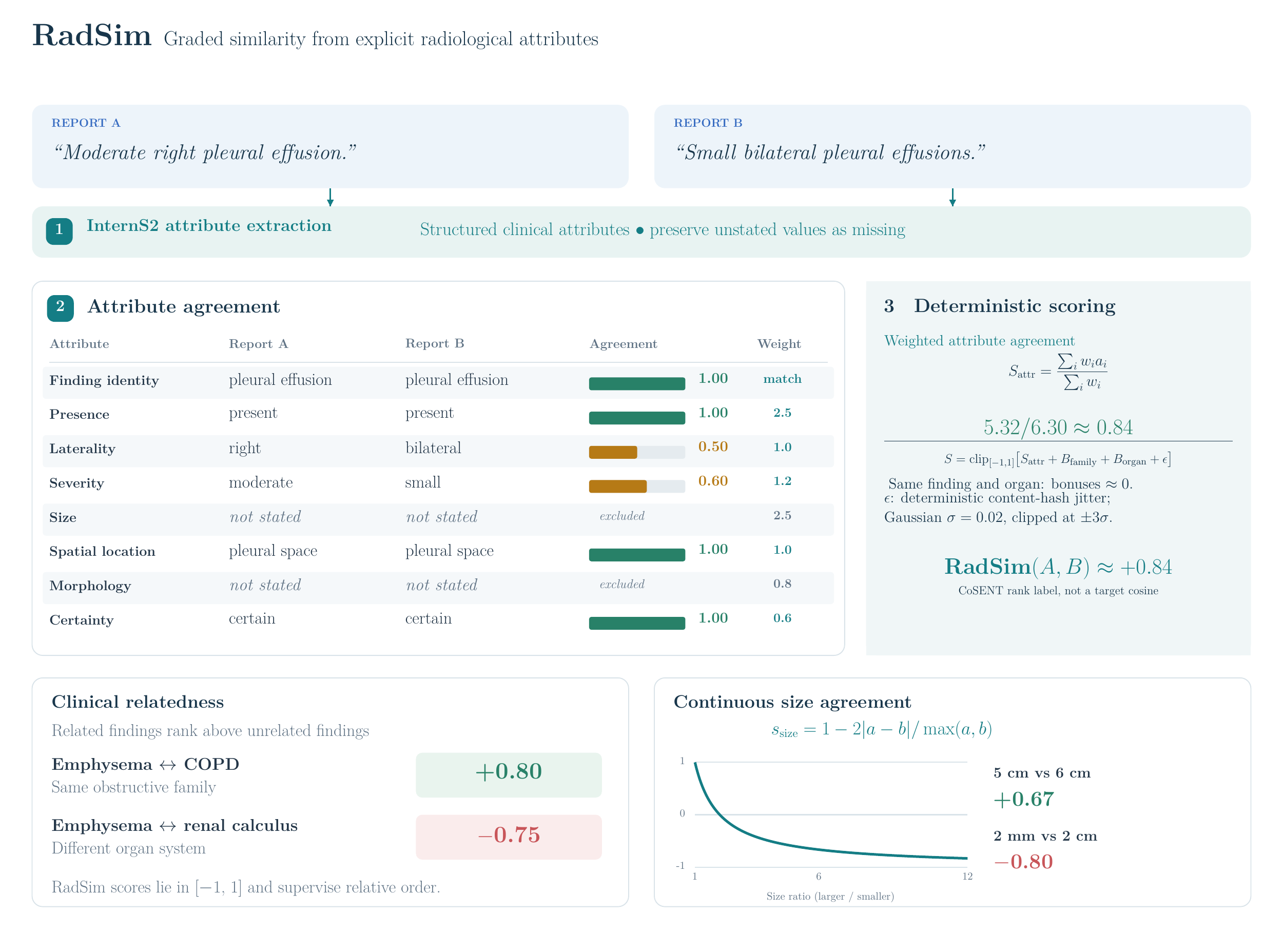}
\caption{RadSim supervision. InternS2 extracts clinical attributes before deterministic scoring. In the example, partial laterality and severity agreement yields $5.32/6.30\approx0.84$; unstated attributes are excluded. The lower panels illustrate disease-family relatedness and continuous size agreement.}
\label{fig:radsim-scoring}
\end{figure}

\subsubsection{Perturbations and order invariance}
Report comparisons should change when findings change, not merely when sentences move. Reordered findings and impressions receive target $1.0$ and comprise $50.2\%$ of the perturbation/reordering set. Adding $n$ findings to a report with $k$ findings uses the tag-Jaccard target $k/(k+n)$; omission uses prominence-weighted retention so that removing a prominent finding matters more than removing an incidental one. The corrupted half of the set has median target $0.65$. These scored pairs are trained with CoSENT, alongside similarity-gap-mined report and sentence triplets.

Hamming error is a complementary \emph{evaluation} of finding-label recovery, not the embedding training loss:
\[
H=\frac{1}{K}\sum_{k=1}^{K}\mathbf{1}[\widehat y_k\ne y_k].
\]
It measures how often recovered presence labels disagree with the target label vector. Separating this diagnostic from cosine response avoids conflating sensitivity to changed content with a calibrated account of additions and omissions.

\subsubsection{Single-attribute contrasts and location control}
Stage~2 samples canonical clinical states and independently realizes anchor, positive, and negative text while changing one attribute. The ten axes are size, count, laterality, lobe location, severity, temporal change, density, margin, certainty, and distribution. Non-target clinical fields are fixed; synonyms, lead-ins, and phrasing vary. Positive--anchor word Jaccard must be below $0.8$, and the target positive--negative gap is at least $0.20$. Exact-string-disjoint splits allocate 90/5/5\% to training/validation/test.

Graded targets use $2\exp[-d^2/(2\sigma^2)]-1$, with $d$ an ordinal distance or log-base-two size/count ratio. Location uses cranio-caudal zone distance with an additional $2.5$ for a side difference. Laterality scores are $1.0$ for matching side, $-0.6$ for left versus right, and $0.15$ for bilateral versus unilateral. Each axis has approximately $45$k--$81$k training triplets; size and count add $22.5$k scored pairs each.

Location-controlled laterality fixes the zone while changing the side, removing the shortcut of answering a laterality comparison through lobe distance. Its $81$k triplets comprise $85\%$ left/right and $15\%$ bilateral/unilateral contrasts. Combined with separate location supervision, this encourages both effects to remain distinguishable without letting stronger location variation overwhelm laterality.

\subsubsection{Joint-attribute priorities}
Stage~3 introduces $50$k pairs over eight combinations: size--laterality, size--location, severity--laterality, severity--location, density--laterality, count--laterality, severity--temporal change, and density--margin. Two attributes vary while all other canonical fields are fixed:

\begin{equation}
y(x,x')=\operatorname{clip}_{[-1,1]}\left[\prod_{a\in\mathcal{G}}\exp\left(-\frac{d_a(x,x')^2}{2\sigma_a^2}\right)-\pi_{\mathrm{lat}}(x,x')\right],
\label{eq:crossaxis}
\end{equation}
where $\mathcal{G}$ contains the graded axes and $\pi_{\mathrm{lat}}$ is $0$ for the same side, $0.5$ for bilateral versus unilateral, and $0.9$ for left versus right. Size/count use log-ratio bandwidth $\sigma=2.5$. For fixed values of the other attributes, subtracting the lateral penalty lowers the target for a side mismatch. Across pairs with different values on other axes, the target reflects the combined graded agreement and laterality penalty. CoSENT learns the resulting ordering, not literal target cosines.

\begin{designintuitionbox}
Learning each attribute separately is insufficient. If a nodule matches in size but is on the wrong side, the size match should not erase the side mismatch. Location-controlled examples isolate the signal; cross-axis examples set its priority when details vary together.
\end{designintuitionbox}

\subsubsection{RadThought and longer-context alignment}
RadThought derives hierarchical representations from approximately $690$k CXR reports: findings become logical evidence predicates, followed by first-, second-, and third-order descriptions and a final synthesis. The predicates record findings, anatomy, and relations \cite{gelfond1988,gelfond1991}; inference requires no symbolic solver. Same-report pairs and findings-alignment, coherence, and derivation triplets connect the hierarchy while contrasting unrelated evidence.

Scene-graph pairs use predicate Jaccard, first-order pairs use RadSim, and higher-order pairs use conclusion attributes including assertion, certainty, severity, acuity, temporal direction, mechanism, and significance. Fused scores weight scene graph, first, second, third, and synthesis levels by $0.10$, $0.30$, $0.22$, $0.18$, and $0.20$. Approximately $135$k fused pairs and $40$k pairs per thought level complement the triplets. The intended benefit is retention of evidence across longer, differently organized descriptions; cross-level retrieval measures alignment, not independent reasoning correctness.

\begin{figure}[!htbp]
\centering
\includegraphics[width=\linewidth]{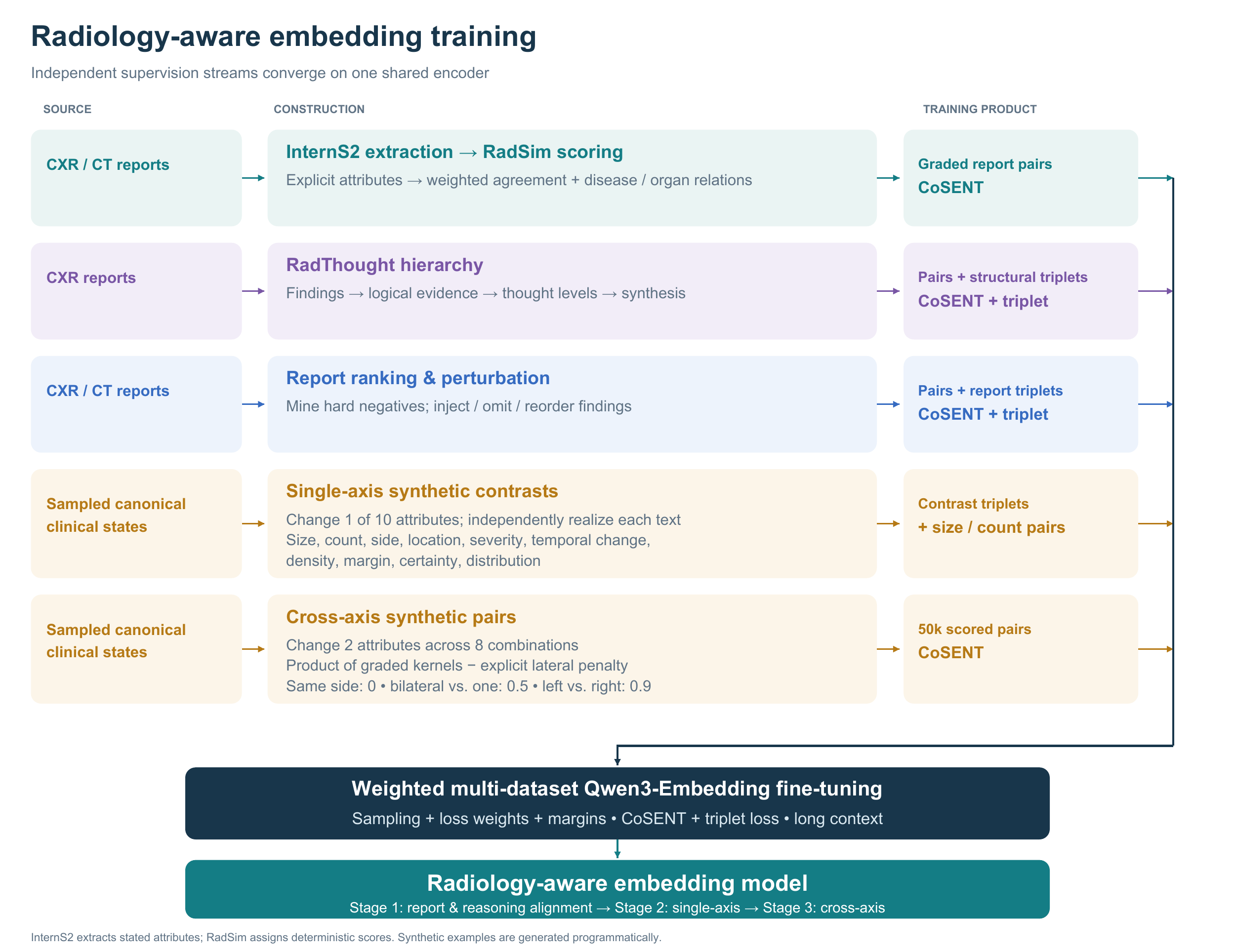}
\caption{Supervision and training pipeline. InternS2 attribute extraction precedes RadSim scoring. Report ranking and perturbation, RadThought hierarchy, single-axis contrasts, and cross-axis pairs supply complementary objectives to one shared encoder.}
\label{fig:pipeline}
\end{figure}

\FloatBarrier
\subsection{Encoder and Learning Objectives}
Qwen3-Embedding-4B is fully fine-tuned. The final-layer end-of-sequence state is normalized to a 2,560-dimensional vector; queries and candidates share the encoder without instruction prefixes. The training sequence-length cap increases from 128 to 9,500 tokens within Stage~1. Long-context training uses eight H100 GPUs, bf16/tf32, gradient checkpointing, cosine learning-rate decay, 3\% warmup, and weight decay $0.01$. Full settings are in Appendix~\ref{app:config}.

\subsubsection{Complementary ranking losses}
For pairs $(u_i,v_i)$ with target scores $s_i$, CoSENT penalizes ranking violations:

\begin{equation}
\mathcal{L}_{\mathrm{CoSENT}}
=
\log\left(
1 + \sum_{(i,j):s_i>s_j}
\exp\left[
\lambda\left(
\cosim(u_j,v_j)-\cosim(u_i,v_i)
\right)
\right]
\right),
\label{eq:cosent}
\end{equation}

The scale is $\lambda=20$. CoSENT does not require target $0.8$ to become cosine $0.8$; it preserves the relative agreement implied by the scoring policy. Triplets instead impose a local separation:
\begin{equation}
\mathcal{L}_{\mathrm{triplet}}=\max(0,s_\theta(a,n)-s_\theta(a,p)+m).
\label{eq:triplet}
\end{equation}
Margins are task specific: too small leaves many triplets inactive, while too large demands an unrealistic gap among highly similar reports. Dataset sampling probabilities $p_i$ and loss weights $w_i$ jointly set the mixture, $\mathcal L=\sum_i p_iw_i\mathbb E[\mathcal L_i]$. Validation losses normalized to each stage's initial value track optimization; retrieval and attribute regression tests guide checkpoint selection.

\subsection{Three-Stage Training Curriculum}
\label{sec:curriculum}
\subsubsection{Stage 1: report and reasoning alignment} RadSim pairs, report/sentence triplets, initial size/laterality contrasts, perturbations, and RadThought are introduced and rebalanced within one stage, including the expansion to long context. This establishes report-level alignment but leaves several controlled attributes weak.

\subsubsection{Stage 2: attribute coverage and rebalancing} Ten-axis contrasts, size/count pairs, and location-controlled laterality broaden sensitivity. Core datasets are capped at 60k examples. Location uses weight $2.0$ and margin $0.30$; controlled laterality uses weight $3.0$ and margin $0.45$ to preserve side discrimination while learning zone differences.

\subsubsection{Stage 3: cross-axis calibration} Joint-attribute pairs address the observation that size, severity, or count can suppress a side mismatch. Cross-axis pairs receive CoSENT weight $1.0$; other datasets are capped at 10k each and single-axis weights become $1.5$. The selected endpoint follows 900 steps at effective batch size 240. Stage trajectories characterize this combined recipe, rather than isolate causal effects of individual components.

\section{Results and Analysis}
\label{sec:results}
\subsection{Evaluation Protocol}
\label{sec:eval}
\paragraph{External report retrieval.}
Open-I contains 3,419 paired studies; testing XR and testing CT each contain 5,000 studies from four institutions outside training. Additional testing pools contain 5,000 CT, 5,000 MR, and 4,276 ultrasound studies, with one study per patient. We label these testing CT-B, testing MR, and testing US to distinguish them from the first testing CT pool. MR and ultrasound were absent from the training corpus and assess transfer to additional modalities. Recall@1 retrieves the source study's paired section in either direction. Open-I also uses label-Jaccard nDCG@10 and 17 short finding queries with precision and contradiction rates at ten results.

\paragraph{Chest retrieval protocol.}
The chest retrieval pool contains 2,336 findings--impression pairs: 467 normal and 1,869 abnormal reports (approximately 20\% and 80\%). Findings and impressions are deduplicated, and findings retrieve their paired impression from the fixed gallery using 128-token inputs. Reports are filtered against training report-triplet anchors, perturbation blocks, and findings-pair text: retained findings blocks are absent from the audited anchors and perturbations, and their sentences are not fully covered by the audited findings-pair pool. These checks define report-level separation. All encoders use the same reports and candidate gallery.

\paragraph{Encoders and comparison scores.}
Ten encoders are evaluated without instruction prefixes: QureRadEmbed, its Qwen3 backbone, zembed-1, BGE-base/large, MedEmbed-base, PubMedBERT-emb, RadEvalModernBERT, BioBERT, and SapBERT. Paired retrieval uses 128 tokens for Open-I, 256 for testing XR, and 512 for testing CT, testing CT-B, testing MR, and testing US; vectors are renormalized in float32. The general validation suite caps Qwen-based encoders at 1,024 tokens and BERT-based encoders at 512. Baseline query instructions are not used, so short-query results describe this common input setting.

Whole-report cosine compares one vector per report. \emph{Align-F1} averages best sentence matches in each direction and takes their harmonic mean. \emph{Directional} takes the minimum of candidate support and abnormal-reference coverage: every candidate sentence must match the reference, while abnormal reference sentences must match the candidate. It uses a keyword/negation rule to identify abnormal sentences and is therefore an embedding-plus-rule method. Sentence embeddings use 128 tokens. GREEN and structured report metrics provide complementary comparisons.

\paragraph{Discrepancies and expert judgment.}
One-sentence edits introduce laterality, presence, severity, size, temporal, omission, or addition errors into external reports. Harmless controls include synonyms, reordering, and Intern-S2 paraphrases/condensations retained after rule and equivalence checks. Win rate is the fraction of paired comparisons in which an erroneous report scores below its harmless rewrite, with half credit for ties. GREEN comparisons use the same 200 source reports per pool for all methods. ReXVal contains 200 candidate pairs across 50 studies annotated by six radiologists; RadEval contains 624 pairs across 208 studies \cite{rexval,xuradeval}. Within-study Kendall $\tau_b$ measures candidate ordering by expert error counts.

\paragraph{Attribute and extraction probes.}
Contrast-10 uses 29,500 held-out synthetic triplets, CrossAxis-8 uses 2,774 pairs, and Probe-13 uses 650 separately written triplets. CrossAxis-8 reports Spearman correlation between cosine similarity and the designed joint-attribute target. Its side-flip response is the mean cosine of same-side pairs minus that of opposite-side pairs; this is a group contrast rather than a matched per-example intervention. Frozen-encoder finding/status probes use 20,000 testing-XR reports (4,000 test) and 3,927 Open-I reports (785 test). Findings plus impression are encoded at 256 tokens. L2-regularized logistic or class-weighted softmax heads predict 14 CheXbert findings/statuses; hyperparameters are chosen on a validation subset of the training portion. Learning curves use 100 through all available training reports, with three repeats at up to 1,000. CheXbert labels are automated supervision; Open-I MeSH labels supply an additional independent label vocabulary.

\FloatBarrier
\subsection{QureRadEmbed Analysis and Open-Source Encoder Comparisons}

\subsubsection{Chest retrieval and curriculum}
\label{sec:chest-retrieval}
Each query is a findings section, and the encoder must rank its paired impression among $2{,}336$ candidates using 128-token inputs. The pool contains 467 normal and 1,869 abnormal reports, with duplicate findings and impressions removed. All encoders and checkpoints are evaluated on the same queries and candidate gallery. The controlled case mix and fixed gallery support direct comparisons within this experiment; absolute recall should not be compared with differently sized or sampled retrieval pools.

\begin{figure}[!htbp]
\centering
\includegraphics[width=\linewidth]{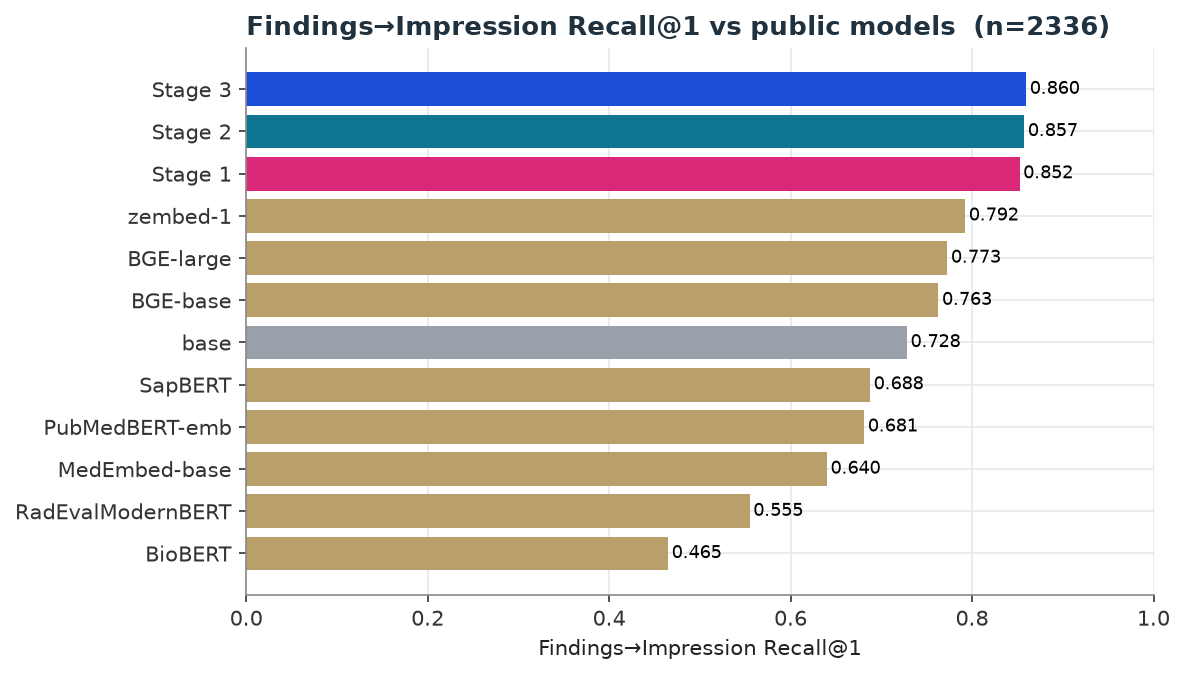}
\caption{Findings-to-impression Recall@1 on the chest retrieval pool (2,336 reports). All three training stages outperform the unadapted backbone and the public encoders evaluated on the same candidate gallery.}
\label{fig:chest-baselines}
\end{figure}

\paragraph{Exact paired-section retrieval.}
Table~\ref{tab:chest-retrieval} reports the complete comparison. Stage~3 places the paired impression first for $86.0\%$ of queries and within the top five and ten for $92.8\%$ and $94.6\%$, respectively, with MRR@10 of $0.890$. Relative to the backbone, the gains are $13.2$, $7.1$, and $5.4$ percentage points at ranks one, five, and ten, and $0.106$ in MRR@10. zembed-1 is the strongest alternative encoder on all five reported metrics. Stage~3 exceeds it by $6.8$, $3.4$, and $3.1$ percentage points in Recall@1/5/10 and by $0.054$ in MRR@10. Thus the improvement extends beyond the first retrieved item. Figure~\ref{fig:chest-baselines} shows the Recall@1 comparison across encoders.

\begin{table}[!htbp]
\centering\small
\caption{Chest findings-to-impression retrieval on the fixed 2,336-report gallery. All values are on a 0--1 scale; higher is better. Abn.~nDCG@10 uses abnormality-label relevance rather than exact study identity. Bold marks the best value in each column across all displayed checkpoints and encoders.}
\label{tab:chest-retrieval}
\setlength{\tabcolsep}{5pt}
\begin{tabular}{lrrrrr}
\toprule
Encoder / checkpoint & R@1 & R@5 & R@10 & MRR@10 & Abn.~nDCG@10\\
\midrule
Qwen3-Embedding-4B & .728 & .857 & .892 & .784 & .234\\
QureRadEmbed Stage 1 & .852 & .922 & .940 & .883 & \textbf{.342}\\
QureRadEmbed Stage 2 & .857 & \textbf{.929} & .944 & .889 & .331\\
QureRadEmbed Stage 3 & \textbf{.860} & .928 & \textbf{.946} & \textbf{.890} & .327\\
\midrule
zembed-1 & .792 & .894 & .915 & .836 & .240\\
BGE-large & .773 & .886 & .914 & .821 & .210\\
BGE-base & .763 & .874 & .908 & .811 & .215\\
SapBERT & .688 & .828 & .878 & .750 & .187\\
PubMedBERT-emb & .681 & .821 & .872 & .743 & .157\\
MedEmbed-base & .640 & .812 & .861 & .715 & .215\\
RadEvalModernBERT & .555 & .715 & .771 & .624 & .131\\
BioBERT & .465 & .609 & .669 & .528 & .149\\
\bottomrule
\end{tabular}
\end{table}

\paragraph{Abnormality relevance beyond study identity.}
Exact retrieval credits only the source study's impression, even when another impression describes similar abnormalities. Abnormality-based nDCG@10 instead evaluates the relevance of the retrieved neighborhood. Stage~3 reaches $0.327$, compared with $0.234$ for the backbone and $0.240$ for zembed-1, absolute gains of $0.093$ and $0.087$. The final model therefore improves both paired-section identification and abnormality-level relevance relative to the alternative encoders. These measures remain distinct: an impression can share the query's abnormality labels without preserving every clinical qualifier or being its paired section.

\begin{figure}[!htbp]
\centering
\includegraphics[width=\linewidth]{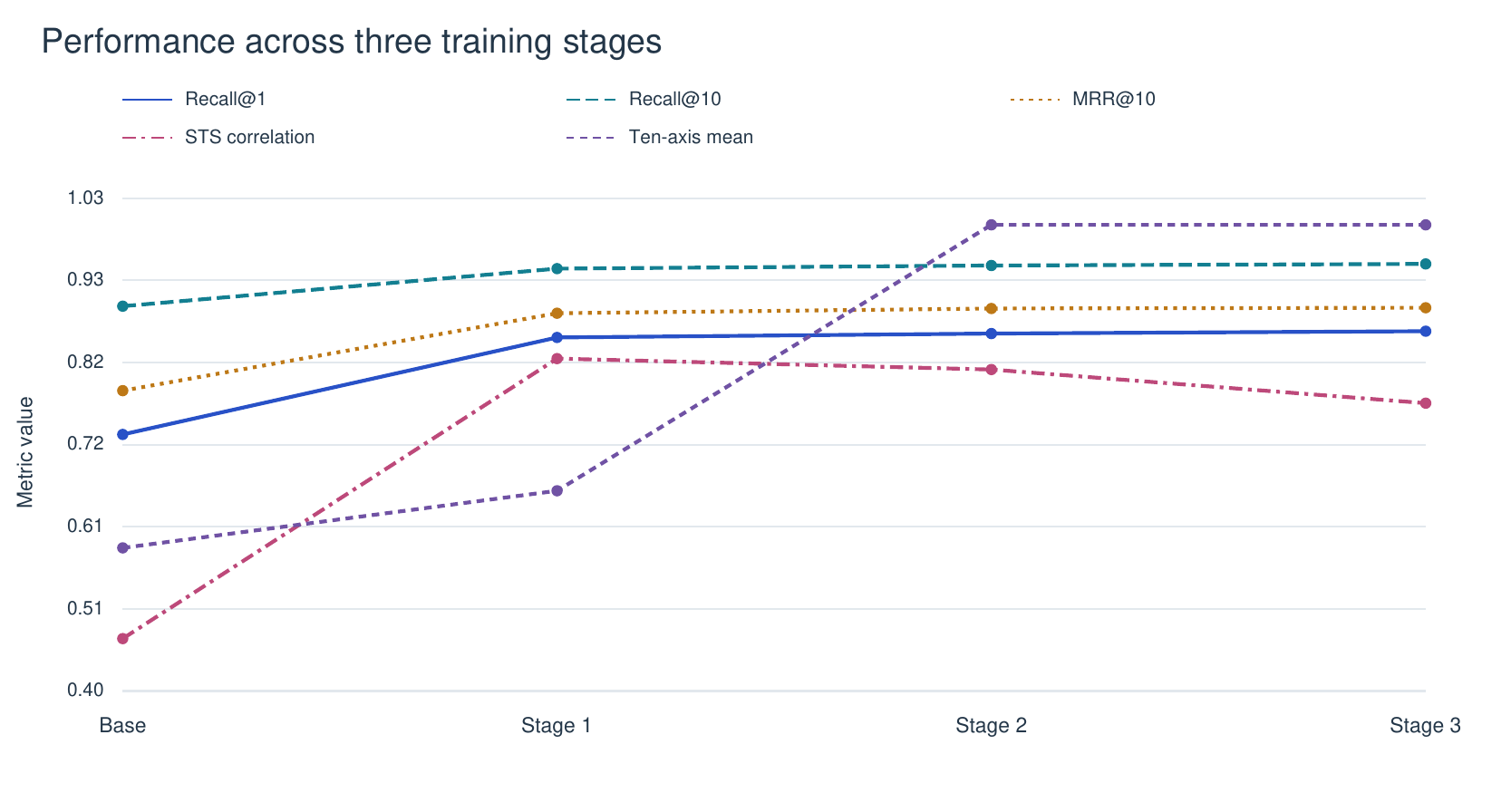}
\caption{Metrics across the base encoder and three training stages. Recall@1, Recall@10, and MRR@10 use the chest retrieval pool (2,336 reports); STS is Spearman correlation, and the ten-axis mean is average contrast-ordering accuracy. Colors and line styles distinguish the metrics.}
\label{fig:curriculum-metrics}
\end{figure}

\paragraph{What changes across the curriculum.}
Stage~1 establishes most of the exact-retrieval improvement: Recall@1 rises from $0.728$ to $0.852$, followed by smaller increases to $0.857$ and $0.860$ at Stages~2 and~3. From Stage~1 to Stage~3, Recall@10 increases from $0.940$ to $0.946$ and MRR@10 from $0.883$ to $0.890$. The final stage is not uniformly better: Recall@5 decreases slightly from $0.929$ to $0.928$, and abnormality nDCG@10 declines from $0.342$ at Stage~1 to $0.331$ and $0.327$. On the separate sentence-similarity benchmark, STS correlation also falls from $0.811$ to $0.768$ between Stages~2 and~3. Figure~\ref{fig:curriculum-metrics} places these trajectories alongside controlled attribute ordering. Together, they show that the later attribute-focused stages retain the earlier retrieval gains while changing the balance between exact matching, broader similarity, and attribute fidelity. The small differences between adjacent checkpoints are descriptive; this evaluation does not report confidence intervals or isolate the effects of individual training components.

\subsubsection{External retrieval and clinically relevant search}
QureRadEmbed ranks first among the ten encoders in both directions on all six external modality pools (Table~\ref{tab:external}). Compared with the strongest alternative in each findings-to-impression task, the gains are $2.8$ percentage points on Open-I, $5.4$ on testing XR, and $11.5$ on testing CT. MR and ultrasound gains show that the learned distinctions transfer beyond the training modalities, rather than only improving matching of familiar chest-report templates.

\begin{table}[!htbp]
\centering\small
\caption{External retrieval Recall@1 (\%). Each cell lists two Recall@1 values separated by a slash: the first is findings-to-impression retrieval, and the second is impression-to-findings retrieval; the slash is a separator, not a ratio. Selected strong comparators are shown; QureRadEmbed leads all ten evaluated encoders on each task.}
\label{tab:external}
\setlength{\tabcolsep}{4pt}
\begin{tabular}{lrrrr}
\toprule
Pool & QureRadEmbed & Qwen3-4B & zembed-1 & BGE-large\\
\midrule
Open-I & \textbf{10.5 / 11.2} & 6.6 / 7.8 & 7.7 / 5.3 & 5.8 / 8.5\\
Testing XR & \textbf{12.4 / 12.9} & 5.7 / 8.7 & 7.0 / 6.6 & 3.5 / 7.3\\
Testing CT & \textbf{42.9 / 45.6} & 31.4 / 39.4 & 28.9 / 33.4 & 9.8 / 24.5\\
Testing CT-B & \textbf{39.9 / 44.1} & 32.0 / 36.7 & 34.4 / 33.8 & 25.0 / 35.2\\
Testing MR & \textbf{54.9 / 60.4} & 47.5 / 52.8 & 44.2 / 47.8 & 31.0 / 44.9\\
Testing US & \textbf{32.9 / 33.4} & 23.6 / 29.9 & 27.1 / 24.7 & 22.0 / 26.9\\
\bottomrule
\end{tabular}
\end{table}

Exact identity is demanding when many normal impressions are interchangeable. On Open-I, graded findings-to-impression nDCG@10 is $0.457$, versus $0.440$ for zembed-1 and $0.368$ for the backbone. More directly relevant to clinical search, short finding-query precision@10 is $0.747$ versus $0.588$ for the backbone. No top-ten result contradicts those 17 finding queries by stating the finding as absent, compared with $17.1\%$ for the backbone. This result indicates that the evaluated model distinguishes assertion status in retrieval; it does not isolate the contribution of presence supervision. Side-specific search remains less consistent: wrong-side@10 is $12.7\%$, versus $3.6\%$ for zembed-1 on the small MeSH-qualified query set. Strong controlled laterality behavior therefore does not guarantee robust side filtering in every retrieval setting.

\subsubsection{Graded attributes and interactions}
The mean Contrast-10 accuracy increases from $0.656$ at Stage~1 to $0.996$ at Stage~2 and stays $0.996$ at Stage~3. All ten final axes exceed $0.98$. Probe-13, whose phrasing is independently written, improves from $0.935$ to $0.965$ mean match accuracy and from $0.816$ to $0.943$ ordinal accuracy between Stages~2 and~3. The agreement between these probes supports learned attribute distinctions beyond the original templates, although density ($0.838$) and certainty ($0.874$) remain harder to match.

\begin{figure}[!htbp]
\centering
\includegraphics[width=\linewidth]{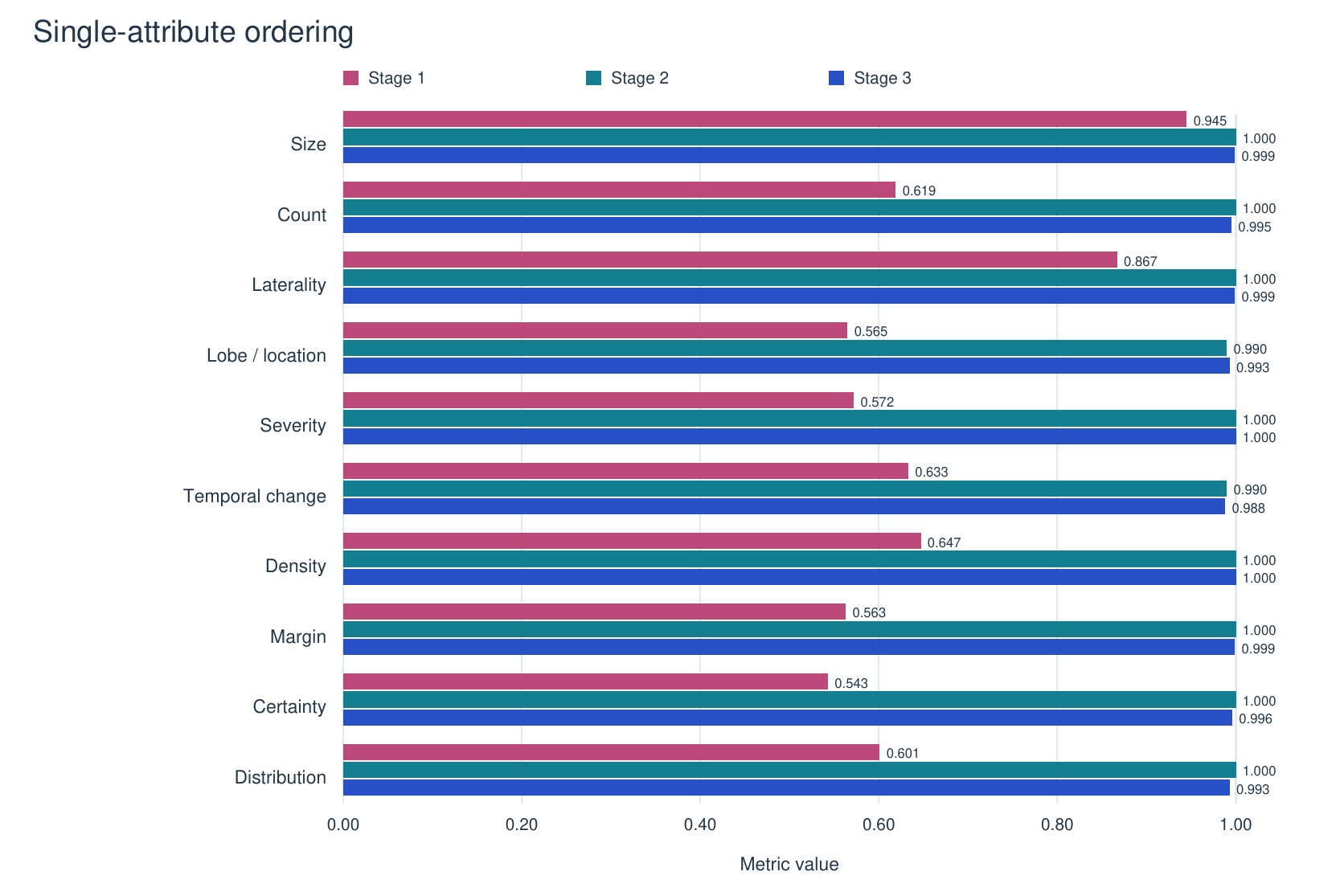}
\caption{Controlled triplet-ordering accuracy across ten attributes. Stage~2 broadens discrimination beyond size and laterality, and Stage~3 retains high accuracy across the expanded suite. Stage~2 per-axis values are reported to two decimal places.}
\label{fig:axes}
\end{figure}

\begin{figure}[!htbp]
\centering
\includegraphics[width=\linewidth]{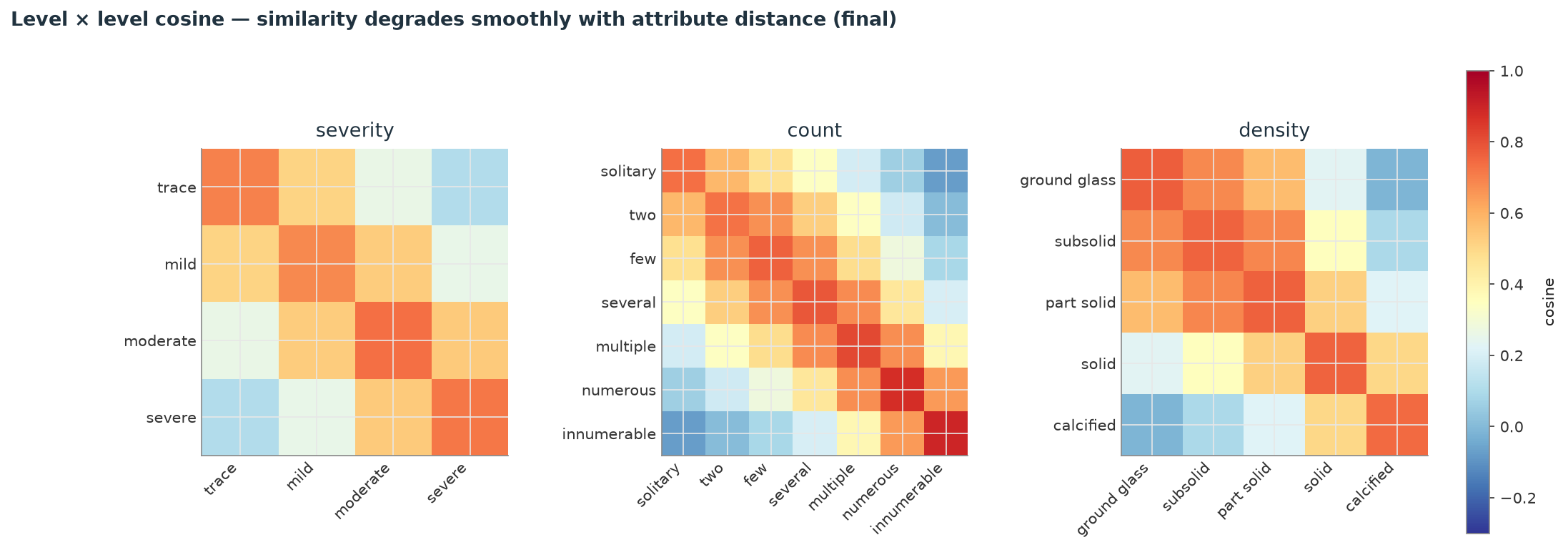}
\caption{Graded attribute structure in the final model. The severity panel spans trace, mild, moderate, and severe; similarity decreases as the descriptions become further separated in severity. Count and density panels show the corresponding cosine structure for their displayed levels. }
\label{fig:axisgeometry}
\end{figure}

\begin{figure}[!htbp]
\centering
\begin{tikzpicture}
\node[anchor=south west,inner sep=0] (plot) at (0,0) {\includegraphics[width=\linewidth]{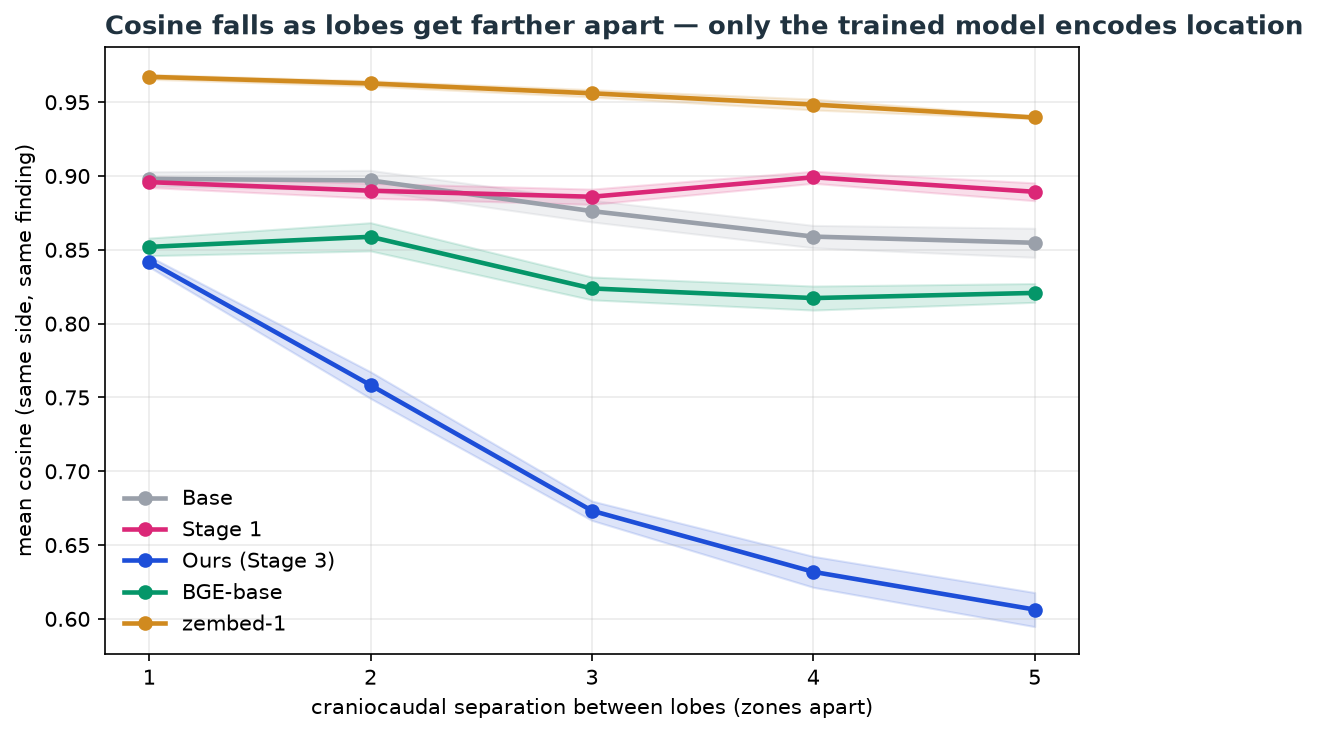}};
\begin{scope}[x={(plot.south east)},y={(plot.north west)}]
\fill[white] (0,0.942) rectangle (1,1);
\node[font=\sffamily\fontsize{8}{9}\selectfont,inner sep=0] at (0.5,0.973) {Similarity across increasing lobe separation};
\end{scope}
\end{tikzpicture}
\caption{Mean cosine similarity versus cranio-caudal location separation for descriptions with the same finding and side. Separation is expressed in ordered zone steps, not physical distance. The final model shows a substantially stronger decrease with increasing separation than the comparison encoders. Shaded bands are descriptive and are not interpreted as confidence intervals.}
\label{fig:lobe-distance}
\end{figure}

The geometry is graded, not merely binary. Severity similarity declines from $0.51$ for trace--mild to $0.26$ for trace--moderate and $0.11$ for trace--severe (Figure~\ref{fig:axisgeometry}). With finding and side fixed, increasing zone separation lowers cosine from approximately $0.84$ to $0.61$ (Figure~\ref{fig:lobe-distance}). These responses follow the intended ordering of attribute differences; their cosine magnitudes are not calibrated measures of clinical importance.

\begin{figure}[!htbp]
\centering
\includegraphics[width=.95\linewidth]{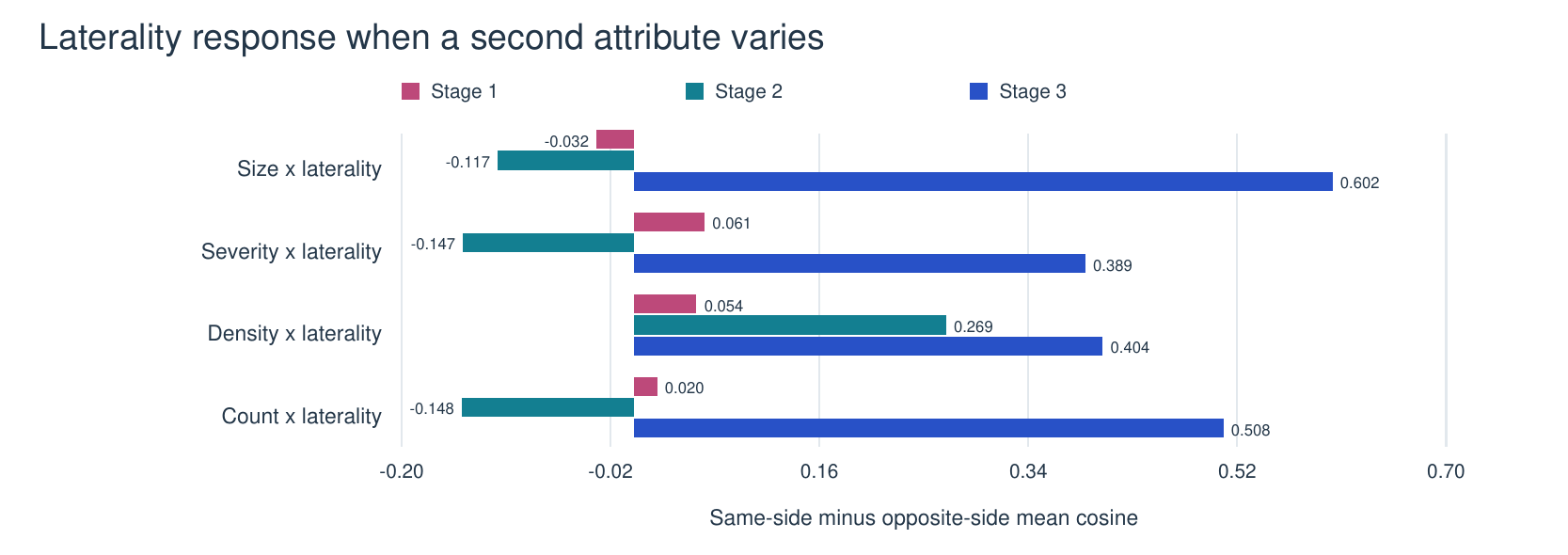}
\caption{Laterality response when another attribute varies, measured as the mean cosine of same-side pairs minus that of opposite-side pairs. Positive values favor the same-side group. The Stage~3 responses are positive for the size, severity, and count combinations that have negative responses at Stage~2.}
\label{fig:cross-response}
\end{figure}

Near-perfect isolated accuracy nevertheless coexists with an interaction failure in the benchmark. At Stage~2, the mean same-side-minus-opposite-side cosine differences for size, severity, and count combinations are negative ($-0.117$, $-0.147$, and $-0.148$). At Stage~3 these differences are positive ($0.602$, $0.389$, and $0.508$), while Spearman correlation with the designed joint-attribute target rises from $0.501$ to $0.976$ (Figure~\ref{fig:cross-response}). These changes are consistent with a different balance of attributes under the Stage~3 recipe; they do not isolate joint supervision from the accompanying reweighting and additional training. The trade-off is visible on the legacy size probe, where accuracy falls from $0.998$ to $0.707$ despite $0.999$ on the newer size generator. No one scalar similarity simultaneously maximizes every attribute diagnostic.

\subsubsection{Clinical organization and frozen-encoder tag extraction}
\begin{figure}[!htbp]
\centering
\includegraphics[width=\linewidth]{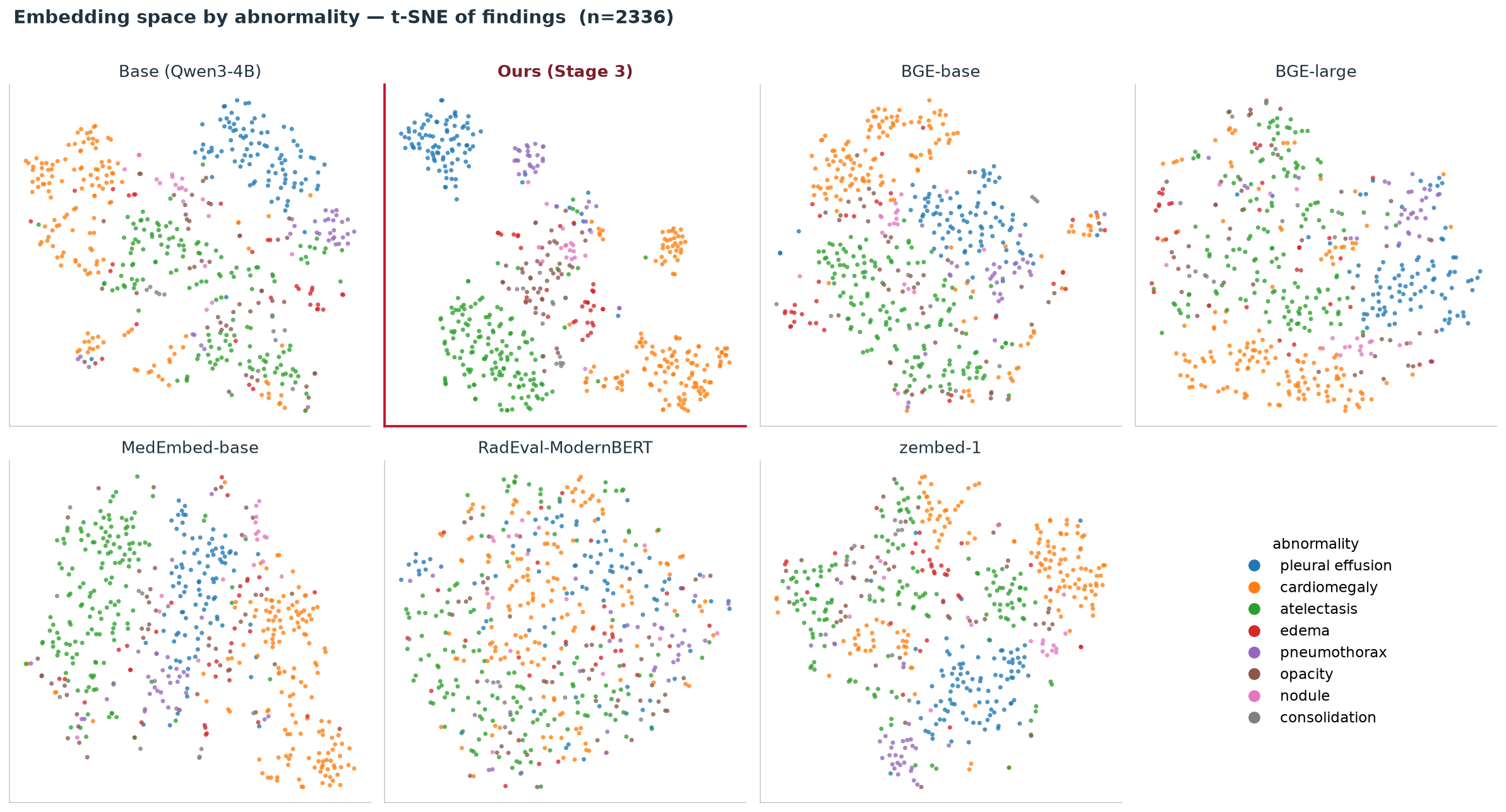}
\caption{t-SNE of findings embeddings from the chest retrieval pool (2,336 reports), shown for the backbone, final QureRadEmbed model, and five public encoders. Colors denote eight abnormalities. The highlighted model separates several clinical groups more clearly; projection distances are not a common metric across panels.}
\label{fig:tsne}
\end{figure}

The t-SNE visualization uses findings from the chest retrieval pool (Figure~\ref{fig:tsne}). QureRadEmbed shows more distinct groupings for several abnormalities, including pleural effusion, cardiomegaly, and pneumothorax, while other groups remain intermixed. This qualitative organization is consistent with the model's clinical similarity objective; a two-dimensional projection alone does not quantify separation in the original embedding space.

\begin{figure}[!htbp]
\centering
\includegraphics[width=\linewidth]{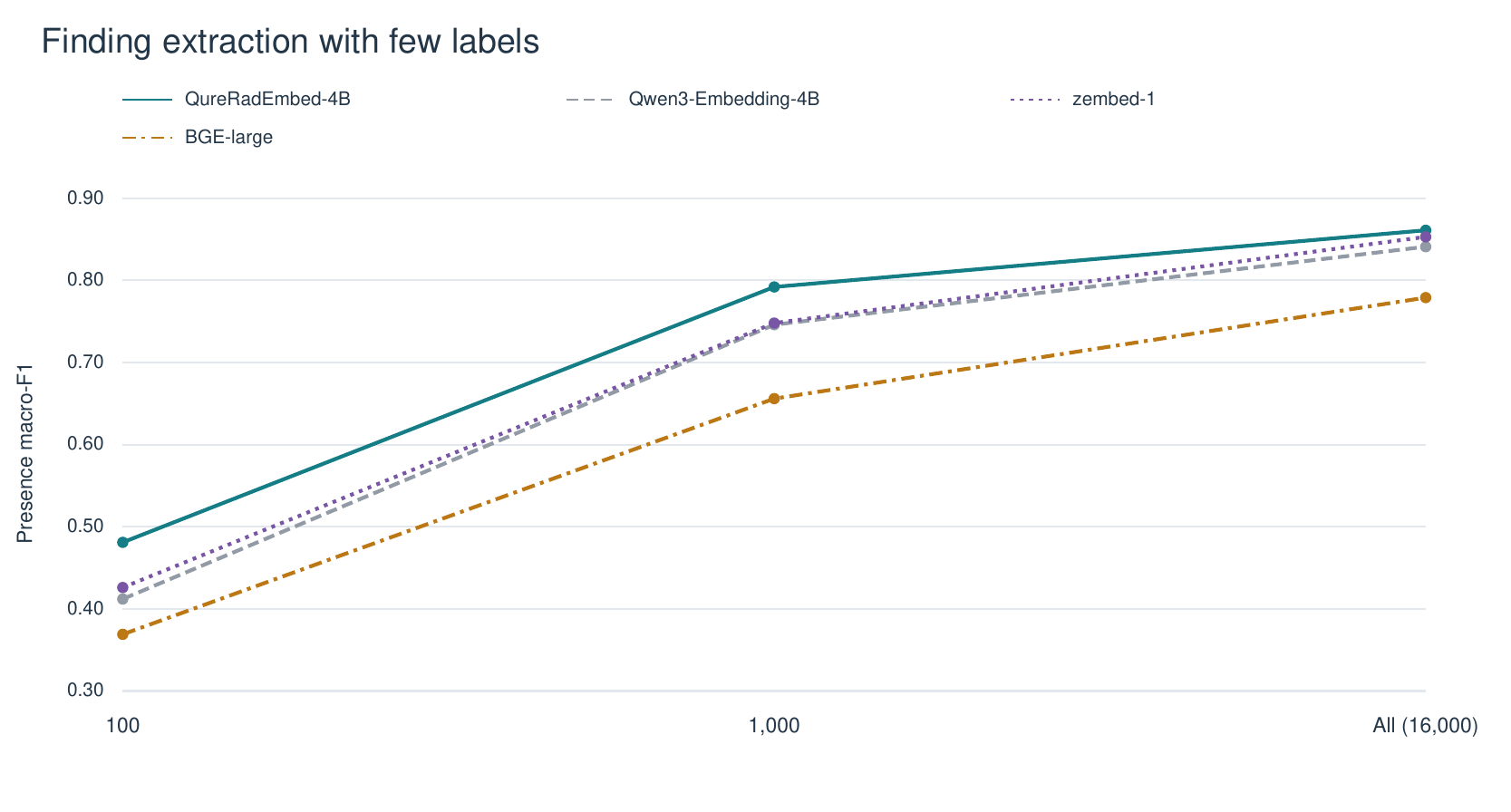}
\caption{Finding-presence linear probes on frozen embeddings. Macro-F1 is measured on 4,000 testing-XR reports; the x-axis is the number of labeled training reports. ``All'' uses the 16,000-report training portion, with internal validation for regularization. Gains are largest in the low-label setting.}
\label{fig:probes}
\end{figure}

Linear probes test whether these representations are useful without retraining the encoder. On testing XR, presence macro-F1 is $0.481$ with 100 labeled reports, $0.792$ with 1,000, and $0.861$ with the full training split (Figure~\ref{fig:probes}). The backbone reaches $0.412$, $0.746$, and $0.841$, respectively; zembed-1 reaches $0.426$, $0.748$, and $0.853$. The larger advantage at 100 labels suggests easier adaptation when a site's labeled data are scarce, rather than only a gain after extensive head training.

Four-status macro-F1 is $0.773$ versus $0.733$ for the backbone, and uncertain-status F1 is $0.449$ versus $0.368$. On Open-I, QureRadEmbed's MeSH-label F1 is $0.878$ versus $0.840$ for the backbone; CheXbert presence F1 is close to zembed-1 ($0.749$ versus $0.753$). These results support report-level tagging for cohort search and structured indexing. They do not establish span extraction, and the status results measure agreement with automated CheXbert labels rather than new radiologist annotations. Uncertainty remains harder than presence, and absent-to-unmentioned confusion reaches $14.9\%$ on Open-I.

\subsubsection{Discrimination across a broad tag vocabulary}
An additional evaluation covers 131 chest X-ray tags in 189,312 test reports and 256 CT tags in 28,645 test reports. Each tag has both positive and negative reference examples. Table~\ref{tab:broad-tag-summary} summarizes these prediction-file results; Appendix~\ref{app:per-tag} reports every tag, including rare and poorly performing tags. The vocabularies include findings, devices, and other report concepts, and should not be interpreted as distinct disease counts.

The supplied scores show strong discrimination across a broad tag vocabulary: macro AUROC is $0.988$ for X-ray and $0.975$ for CT, and AUROC is at least $0.90$ for 130 of 131 X-ray tags and 243 of 256 CT tags. These threshold-independent ranking results do not establish superiority over another labeling model, because a matched labeling baseline is not included in this experiment.

For an exploratory operating-point analysis, each tag uses the threshold maximizing Youden's $J=\mathrm{sensitivity}+\mathrm{specificity}-1$, with scores at or above the threshold classified as positive. Thresholds are selected on the same test examples used to compute the operating-point metrics; these are therefore test-selected estimates rather than performance at independently validated thresholds. Macro metrics weight each tag equally.

At these thresholds, macro sensitivity/specificity are $0.967/0.959$ for X-ray and $0.940/0.938$ for CT, with macro-F1 of $0.350$ and $0.301$, respectively. Thus, strong score discrimination coexists with heterogeneous binary-label performance. Validation-selected thresholds and independent evaluation are needed to establish reliable automatic labeling, particularly for rare tags.

\noindent\textbf{Protocol qualification.} These additional results were computed from supplied reference labels and continuous scores, recomputing all binary decisions at the tag-specific maximum-Youden thresholds. The generating checkpoint and labeling head, encoder freezing or fine-tuning, and reference-label provenance have not yet been documented for this experiment. Consequently, these results are provisional and are not treated as evidence of frozen-encoder transfer or gains attributable to the proposed supervision. These details and matched-backbone comparisons are required before making either claim.

\begin{table}[!htbp]
\centering
\caption{Discrimination across a broad tag vocabulary and exploratory operating-point results. AUROC uses continuous scores. Sensitivity, specificity, and F1 use tag-specific maximum-Youden thresholds selected on the evaluated test examples. Macro metrics weight tags equally.}
\label{tab:broad-tag-summary}
\begin{tabular}{lrr}
\toprule
Metric & X-ray & CT \\
\midrule
Test reports & 189,312 & 28,645 \\
Tags & 131 & 256 \\
Macro AUROC & 0.988 & 0.975 \\
Tags with AUROC $\geq 0.90$ & 130 / 131 & 243 / 256 \\
Macro sensitivity & 0.967 & 0.940 \\
Macro specificity & 0.959 & 0.938 \\
Macro Youden $J$ & 0.925 & 0.879 \\
Macro F1 & 0.350 & 0.301 \\
\bottomrule
\end{tabular}
\end{table}

\subsubsection{Attribute-specific similarity from frozen embeddings}
To test whether one embedding exposes distinct clinical attributes, we train 11 linear projection heads on cached, frozen Stage~3 embeddings. Each head maps a report or finding vector to 128 dimensions and computes
\[
s_a(x,y)=\cosim\!\left(W_a f_\theta(x),W_a f_\theta(y)\right).
\]
Cosine mean-squared error fits the graded target for the selected attribute; pairs changing only other attributes receive target $1.0$, encouraging invariance. Training combines single-axis, presence, and two-axis pairs, with weaker axes up-weighted. The configuration uses batch size 512, learning rate $10^{-3}$, and 25--40 epochs on one L40S. Evaluation uses 10,450 pairs (950 per attribute) and a separate 2,774-pair cross-axis set.

The heads achieve mean own-attribute Spearman correlation $0.940$ and mean AUC $0.994$; the presence head distinguishes present--absent contradictions with AUC $1.000$. The diagonal-to-off-diagonal absolute-correlation ratio is $3.634$, indicating that the heads respond more strongly to their intended attributes than to other changes. These results support learning attribute-specific similarity through linear projections of frozen embeddings, beyond the report-level finding/status classification above. The projection is linear, while the subsequent cosine score is not a linear function of the original embedding pair.

\begin{table}[!htbp]
\centering\small
\caption{Aggregate frozen-encoder attribute-head results. Modularity is the ratio of intended-attribute response to off-attribute response. The remaining measures characterize separation and concentration of attribute information; these do not establish orthogonal subspaces.}
\label{tab:attribute-heads}
\begin{tabular}{lr}
\toprule
Metric & Value\\
\midrule
Mean own-attribute Spearman $\rho$ & 0.940\\
Mean AUC & 0.994\\
Contradiction AUC & 1.000\\
Modularity ratio & 3.634\\
DCI disentanglement & 0.073\\
DCI completeness & 0.076\\
Mutual information gap (MIG) & 0.165\\
Separated attribute predictability (SAP) & 0.553\\
\bottomrule
\end{tabular}
\end{table}

The distinction between readout and full disentanglement matters. High attribute prediction with modest DCI scores means useful signals can be extracted without proving that each attribute occupies an independent, orthogonal subspace. Clinically, the heads offer a way to ask whether two descriptions agree on a particular detail rather than compress all disagreement into one cosine. For example, separate size and laterality scores could expose a size match on the wrong side. Performance of such a combined decision rule remains a separate evaluation.

\FloatBarrier
\subsection{Open-Source Evaluation Metrics and Comparisons}
\subsubsection{Report discrepancies: local matching matters}
On the full Open-I discrepancy set, whole-report cosine ranks major errors below synonym controls with mean win rate $0.907$, compared with $0.574$ for the backbone and $0.803$ for zembed-1. Presence changes are particularly clear: negation, affirmation, omission, and addition win rates are $0.991$, $0.999$, $0.998$, and $0.997$. This complements the retrieval result: the model responds to changed assertions rather than only matching finding vocabulary. Omission and addition resemble training perturbations, so the other edit types provide a separate test of transfer.

Real paraphrases are harder than local synonym replacements (Table~\ref{tab:discrepancy}). Shared report content can hide a one-sentence error, while broad rewording moves the whole-report vector. On testing XR, directional sentence scoring improves the error-versus-paraphrase win rate from $0.537$ to $0.618$; on testing CT it improves from $0.541$ to $0.717$. GREEN reaches $0.743$ and $0.783$. For side flips specifically, cosine scores only $0.097$ and $0.074$, versus $0.452$ and $0.630$ for directional scoring and $0.806$ and $0.778$ for GREEN. Thus local matching mitigates dilution but does not eliminate the hard cases.

\begin{table}[!htbp]
\centering\small
\caption{Discrepancy win rates on common GREEN subsets: 1,356 Open-I, 1,230 testing-XR, and 1,374 testing-CT pairs. P compares an erroneous report with an LLM paraphrase; C compares it with an LLM condensation. Each value is the fraction of comparisons in which the error receives a lower similarity score than the compatible rewrite (ties receive half credit). Higher is better.}
\label{tab:discrepancy}
\setlength{\tabcolsep}{6pt}
\begin{tabular}{lrrrrrr}
\toprule
& \multicolumn{2}{c}{Open-I} & \multicolumn{2}{c}{Testing XR} & \multicolumn{2}{c}{Testing CT}\\
\cmidrule(lr){2-3}\cmidrule(lr){4-5}\cmidrule(lr){6-7}
Method & P & C & P & C & P & C\\
\midrule
QureRadEmbed cosine & .630 & .721 & .537 & .436 & .541 & .358\\
QureRadEmbed directional & .864 & .928 & .618 & .593 & .717 & .499\\
GREEN & .825 & .538 & .743 & .361 & .783 & .433\\
\bottomrule
\end{tabular}
\end{table}

Directional coverage also better tolerates condensation because it does not demand that every normal reference sentence be repeated. Against LLM condensations it scores $0.928$, $0.593$, and $0.499$, compared with GREEN's $0.538$, $0.361$, and $0.433$. This is a task-specific benefit of the coverage rule, not a general property of the encoder; dropping a pertinent negative can still matter clinically. Its weaker testing-CT result also cautions against transferring a chest-oriented abnormal-sentence rule unchanged to longer multi-organ reports.

\subsubsection{Agreement with clinical judgments}
Table~\ref{tab:public-human} summarizes the public human-judgment benchmarks alongside embedding and report-quality metrics.

\begin{table}[!htbp]
\centering\small
\caption{Agreement with human judgments on public benchmarks. Values are Kendall $\tau$ (higher is better). ReXVal uses within-study agreement with clinically significant error counts (200 pairs); RadEval uses within-study agreement with reader error counts (624 pairs). RaTE-Eval evaluates sentence (441 pairs) and paragraph (370 pairs) ratings. The two RaTE columns are separate tasks, not retrieval directions. Selected embedding and report-metric comparators are shown with their scoring rule fixed across columns.}
\label{tab:public-human}
\setlength{\tabcolsep}{5pt}
\begin{tabular}{lrrrr}
\toprule
& ReXVal & RadEval & \multicolumn{2}{c}{RaTE-Eval}\\
\cmidrule(lr){4-5}
Method & Within study & Within study & Sentence & Paragraph\\
\midrule
QureRadEmbed cosine & 0.336 & 0.122 & 0.246 & 0.494\\
QureRadEmbed Align-F1 & 0.301 & 0.178 & 0.245 & 0.384\\
Qwen3-4B cosine & 0.207 & 0.071 & 0.428 & 0.575\\
zembed-1 cosine & 0.217 & 0.174 & 0.298 & 0.588\\
BGE-base cosine & 0.378 & 0.100 & 0.351 & 0.489\\
MedEmbed-base cosine & 0.392 & 0.156 & 0.385 & 0.499\\
SapBERT cosine & 0.229 & 0.082 & 0.437 & 0.366\\
\midrule
BERTScore & 0.227 & 0.149 & 0.285 & 0.474\\
RadGraph F1 & 0.399 & 0.043 & 0.341 & 0.342\\
RaTEScore & 0.223 & 0.130 & 0.352 & 0.462\\
RadCliQ-v1 & 0.414 & 0.103 & 0.299 & 0.481\\
F1-RadBERT-CT & 0.087 & 0.322 & 0.228 & 0.200\\
GREEN & 0.445 & 0.211 & 0.372 & 0.394\\
\bottomrule
\end{tabular}
\end{table}

On ReXVal, whole-report cosine improves within-study agreement with clinically significant error counts from $\tau_b=0.207$ for the backbone to $0.336$ for QureRadEmbed, but remains below GREEN ($0.445$) and RadGraph F1 ($0.399$). On the RadEval reader study, sentence alignment improves QureRadEmbed from $0.122$ to $0.178$; GREEN reaches $0.211$ and F1-RadBERT-CT $0.322$. RaTE-Eval likewise distinguishes retrieval from human-rated similarity: QureRadEmbed cosine has sentence/paragraph correlations of $0.246/0.494$, versus $0.428/0.575$ for its backbone \cite{ratescore}. Attribute-aware adaptation therefore provides useful comparison signals, but does not make cosine a replacement for expert-style error assessment.

Two additional tests clarify what it captures. Cosine separates entailment from contradiction with AUROC $0.958$ on RadNLI and $0.888$ on MedNLI, compared with $0.897$ and $0.665$ for the backbone. This is stronger assertion compatibility, not directional inference: a trained three-class classifier does not improve over the backbone. On the 2,448-report SSREE graded-corruption test, QureRadEmbed has the highest whole-report cosine monotonicity ($\tau_b=0.743$), while sentence alignment reaches $0.777$ and BGE-large Align-F1 reaches $0.841$ \cite{radsem}. The common lesson is that the comparison rule matters alongside the encoder.

\FloatBarrier
\subsection{Inference cost and reuse}
\label{sec:efficiency}
On one NVIDIA L40S, QureRadEmbed processes 226 reports/s at a 256-token cap, with 11.8\,GB peak GPU memory and 5,120 bytes per fp16 embedding. Encoding both texts gives 8.8 seconds per 1,000 whole-report comparisons. GREEN takes 2,755.1 seconds, RaTEScore 99.6, and RadGraph F1 24.5 in the benchmark runs. BGE-base is faster still (1.4 seconds), so QureRadEmbed occupies a useful middle ground between smaller general encoders and generative evaluation.

\begin{table}[!htbp]
\centering\small
\caption{Measured report-comparison cost on one NVIDIA L40S. Values are seconds per 1,000 report pairs (lower is better). Cosine rows include encoding both reports, without cached vectors. Report-metric timings are normalized from the Open-I discrepancy runs; GREEN used a 200-report subset and generation batch size eight. These are implementation-specific measurements, not matched-throughput limits.}
\label{tab:metric-runtime}
\begin{tabular}{lr}
\toprule
Method & Seconds per 1,000 pairs\\
\midrule
BGE-base cosine & 1.4\\
BERTScore & 1.4\\
RadEvalModernBERT cosine & 1.7\\
F1-RadBERT-CT & 1.9\\
BGE-large cosine & 3.7\\
QureRadEmbed cosine & 8.8\\
RadGraph F1 & 24.5\\
RadCliQ-v1 & 26.9\\
RaTEScore & 99.6\\
GREEN & 2,755.1\\
\bottomrule
\end{tabular}
\end{table}

Table~\ref{tab:metric-runtime} compares the measured costs. The approximately $313\times$ GREEN/cosine ratio describes the measured implementations: GREEN used Hugging Face generation with batch size eight. Optimized generation can narrow the gap. Sentence-level directional scoring embeds additional sentences and has no separate end-to-end timing here, so this speedup is not assigned to it. Cached reference vectors reduce repeated encoding further; pairwise cosine and linear tag heads can reuse the same representation across search and comparison workflows.

\section{Discussion and Future Work}
The results identify three complementary uses. \textbf{Clinical retrieval} benefits from fewer negation contradictions and improved external paired-section matching. \textbf{Structured report indexing} benefits from linearly accessible finding/status information, particularly with few labels. \textbf{Report comparison} benefits from sensitivity to altered findings, with sentence-level aggregation retaining details that one pooled vector can dilute. A practical deployment could use embeddings for broad search and inexpensive comparison, then apply structured checks or generative review to selected cases; this workflow remains to be evaluated prospectively.

The results are consistent with task-specific benefits and trade-offs under the attribute curriculum; the checkpoint comparisons do not isolate the effects of individual components. Single-axis contrasts teach that a detail matters; location control separates side from lobe confounding; joint-axis targets specify the intended ordering when other details compete. The final stage's cross-axis gain and legacy size regression expose this trade-off. The attribute heads demonstrate that separate similarity signals can be read from frozen embeddings; combining them with application-specific constraints may offer more control than a single global cosine.

\paragraph{Normal-report ambiguity and abnormality burden.}
Many normal impressions correctly match the same normal findings section, making exact source-study retrieval intrinsically ambiguous. A single abnormality embedded in otherwise normal text offers limited distinguishing information. Multiple abnormalities provide richer combinations of findings and attributes, making radiology-specific similarity more useful, while also increasing the need to retain each local detail. Relevance-aware evaluation should therefore complement exact identity and distinguish normal, single-abnormality, and multi-abnormality cases.

\paragraph{Long context, direction, and clinical importance.}
A longer input limit reduces truncation but does not guarantee that every finding survives pooling. The side-flip/paraphrase results show this remaining gap despite strong phrase-level tests. Hierarchical or finding-level comparison is a natural extension of RadThought. Cosine is also symmetric: omission and addition can have different clinical implications, and a score drop is not a calibrated estimate of harm. Directional support/coverage helps, but its abnormality rule and treatment of pertinent negatives need adaptation across modalities. Future work should combine local attribute checks, broader long-report supervision, and expert evaluation of clinically consequential disagreements.

\section{Conclusion}
QureRadEmbed combines inspectable RadSim targets, hierarchical RadThought supervision, and controlled attribute interventions to learn radiology-aware representations. External retrieval and low-label tagging demonstrate practical value beyond broad biomedical similarity, while joint-axis tests show why individual attribute sensitivity must be paired with appropriate relative priorities. The model provides efficient, reusable clinical comparison signals; local aggregation and generative evaluation remain complementary for detailed report-quality assessment.

\appendix
\section{Additional Report Diagnostics}\label{app:historical}
\paragraph{Hierarchical alignment and perturbations.}
On the report-derived hierarchy, first-to-second-order retrieval rises from $0.365$ for the backbone to $0.639$ at Stage~3; first-order-to-synthesis rises from $0.219$ to $0.666$. These are alignment diagnostics on report-derived text, not independent reasoning benchmarks. In the 300-report injection probe, adding three false findings lowers final-model cosine by $0.216$ versus $0.140$ for the backbone. Best-threshold Hamming label-recovery error is $0.069$ for the backbone, $0.048$ at Stage~1, and $0.050$ at Stage~3. Stronger cosine response and minimum label-recovery error are different objectives.

\paragraph{Larger testing pools.}
\label{sec:ct}
An earlier, separate protocol uses 35,613 testing-XR and 90,951 testing-CT/CTA reports, with findings/impression caps of 384/256 tokens. Backbone-to-final Recall@1 improves from $0.054$ to $0.135$ for XR and $0.187$ to $0.286$ for CT. Stage~2 reaches $0.147$/$0.317$. These results characterize the final stage's trade-off in larger galleries; they are not directly comparable with the 5,000-report, differently truncated evaluations in Table~\ref{tab:external}.

\section{Training Configuration}\label{app:config}
\begin{table}[H]
\centering
\caption{Documented configuration for Stage~1. Later refinements are described in Section~\ref{sec:curriculum}.}
\small
\begin{tabularx}{\textwidth}{l X}
\toprule
Item & Configuration \\
\midrule
Backbones & Qwen3-Embedding-4B, full fine-tuning, no LoRA. \\
Framework & sentence-transformers; plain DDP via \texttt{torchrun}. \\
Precision & bf16 + tf32. \\
Schedule & Cosine learning-rate schedule, 3\% warmup, weight decay 0.01, seed 42. \\
Sequence length & 128 in early development; 9,500 during the long-context portion of Stage~1 and throughout Stages~2--3. \\
Effective batch & 256 during long-context Stage~1 training; 240 in Stages~2--3. \\
Learning rate & $3\times10^{-6}$ in early training; $10^{-5}$ in long-context stages. \\
Losses & CoSENT ($\lambda=20$) for scored pairs; cosine triplet loss for contrast sets; per-dataset static loss weighting. \\
Stage 1 margins & Report and laterality margin 0.45; RadThought margins adjusted by thought-transition difficulty; overall range 0.1--0.5. \\
Stage 1 weights & Report triplet 5.0; RadThought$\rightarrow$findings 4.5; laterality 4.0; size and RadThought coherence/thought-transition 3.5; per-level RadThought pairs 2.0; findings/impression pairs 1.5; perturbation 0.5; sentence pairs 0.05. \\
Gradient checkpointing & Enabled and required at sequence length 9,500. \\
Checkpoint selection & Task-regression suite for checkpoint selection; normalized $\etl$ for monitoring. \\
Labeling engine & Intern-S2-Preview; deterministic RadSim score conversion. \\
Hardware & Eight H100 GPUs for long-context 4B stages. \\
\bottomrule
\end{tabularx}
\end{table}

\subsection{Stage 2 and Stage 3 settings}
Stage~2 uses single-axis weights of $3.0$ for size and location-controlled laterality, $2.5$ for count, $2.0$ for location, severity, temporal change, density, and distribution, and $1.5$ for margin and certainty. Margins are $0.35$, except $0.45$ for laterality and $0.30$ for location. Stage~3 retains the core report/reasoning weights but reduces all single-axis weights to $1.5$.

\begin{table}[!htbp]
\centering\small
\caption{Stage~3 training mixture. Each batch comes from one dataset, sampled in proportion to its capped size.}
\begin{tabular}{lrrr}
\toprule
Dataset & Samples & Weight & Margin\\
\midrule
Cross-axis pairs & 50k & 1.0 & --\\
Findings / impression pairs & 10k each & 1.5 & --\\
RadThought fused pairs & 10k & 0.6 & --\\
RadThought per-level pairs & 10k each & 2.0 & --\\
Perturbation pairs & 10k & 0.5 & --\\
Size / count scored pairs & 10k each & 0.4 & --\\
Report triplets & 10k & 5.0 & 0.45\\
RadThought--findings triplets & 10k & 4.5 & 0.50\\
Coherence / derivation triplets & 10k each & 3.5 & 0.50\\
Sentence triplets & 10k & 0.5 & 0.10\\
Single-axis triplets (ten axes) & 10k each & 1.5 & 0.30--0.45\\
\bottomrule
\end{tabular}
\end{table}

\clearpage
\section{Per-Tag Report Labeling Results}\label{app:per-tag}
Tables~\ref{tab:xray-per-tag} and~\ref{tab:ct-per-tag} list all 131 X-ray and 256 CT tags, respectively, in alphabetical order. Support is the number of reference-positive reports. Sensitivity (Sens.), specificity (Spec.), and F1 are recomputed from continuous scores at the threshold maximizing Youden's $J=\mathrm{Sens.}+\mathrm{Spec.}-1$ for each tag. A score equal to the threshold is classified as positive. All distinct observed score cutoffs and the all-negative endpoint are considered; ties in maximum $J$ are resolved by choosing the highest threshold. Tied scores are handled as a group. AUROC is threshold-independent.

\noindent\textbf{Exploratory threshold selection.} The thresholds are selected using the same test reference labels used for these tables. Operating-point metrics are therefore optimistically selected and must not be interpreted as performance at independently validated thresholds. Thresholds are displayed to six decimal places and metrics to three decimal places; calculations use the full stored score precision. No tag is excluded based on support or performance. Four X-ray tags and 36 CT tags have fewer than 25 positive references. The protocol qualification in the broad-vocabulary tagging results applies to both tables.

\begingroup
\small\setlength{\tabcolsep}{2pt}\renewcommand{\arraystretch}{1.10}
\begin{longtable}{@{}>{\raggedright\arraybackslash}p{5.4cm}rrrrrr@{}}
\caption{Chest X-ray per-tag results (131 tags) at test-selected maximum-Youden thresholds. Support is reference-positive count; Sens. and Spec. denote sensitivity and specificity.}\label{tab:xray-per-tag}\\
\toprule
Tag & Support & Threshold & Sens. & Spec. & F1 & AUROC \\
\midrule
\endfirsthead
\multicolumn{7}{l}{\textit{Chest X-ray: test-selected maximum-Youden thresholds (continued)}}\\
\toprule
Tag & Support & Threshold & Sens. & Spec. & F1 & AUROC \\
\midrule
\endhead
\midrule
\multicolumn{7}{r}{\textit{Continued on next page}}\\
\endfoot
\bottomrule
\endlastfoot
aeration & 414 & 0.039386 & 0.667 & 0.940 & 0.046 & 0.847 \\
airspace\_\allowbreak opacity & 1,522 & 0.308242 & 0.929 & 0.919 & 0.155 & 0.974 \\
aneurysm & 33 & 0.038370 & 1.000 & 0.977 & 0.015 & 0.998 \\
aortic\_\allowbreak abnormality & 1,101 & 0.153177 & 0.942 & 0.934 & 0.142 & 0.975 \\
aortic\_\allowbreak aneurysm & 695 & 0.157237 & 0.944 & 0.948 & 0.117 & 0.990 \\
aortic\_\allowbreak ectasia & 785 & 0.132587 & 0.995 & 0.989 & 0.421 & 0.999 \\
aortic\_\allowbreak tortuosity & 10,626 & 0.367768 & 0.995 & 0.961 & 0.749 & 0.989 \\
ards & 185 & 0.051807 & 0.957 & 0.957 & 0.042 & 0.992 \\
artifact & 13,762 & 0.440248 & 0.869 & 0.895 & 0.542 & 0.954 \\
aspiration & 784 & 0.180101 & 0.968 & 0.963 & 0.177 & 0.994 \\
atelectasis & 33,382 & 0.554103 & 0.978 & 0.989 & 0.965 & 0.998 \\
atherosclerosis & 1,870 & 0.607955 & 0.969 & 0.917 & 0.189 & 0.967 \\
basilar\_\allowbreak opacity & 1,380 & 0.162851 & 0.958 & 0.930 & 0.166 & 0.985 \\
bone\_\allowbreak metastasis & 39 & 0.044306 & 0.974 & 0.967 & 0.012 & 0.984 \\
bronchiectasis & 247 & 0.082162 & 0.951 & 0.990 & 0.199 & 0.994 \\
bronchitis & 2,628 & 0.243714 & 0.941 & 0.961 & 0.397 & 0.990 \\
cabg & 2,864 & 0.206940 & 0.936 & 0.933 & 0.296 & 0.977 \\
calcification & 22,735 & 0.718659 & 0.988 & 0.963 & 0.874 & 0.992 \\
calcified\_\allowbreak granuloma & 250 & 0.226717 & 0.992 & 0.960 & 0.062 & 0.985 \\
calcified\_\allowbreak pulmonary\_\allowbreak nodule & 832 & 0.241835 & 0.989 & 0.959 & 0.177 & 0.986 \\
cardiac\_\allowbreak defibrillator & 402 & 0.269151 & 0.955 & 0.955 & 0.083 & 0.988 \\
cardiomegaly & 34,131 & 0.667891 & 0.986 & 0.981 & 0.951 & 0.994 \\
catheter & 3,241 & 0.416833 & 0.921 & 0.861 & 0.186 & 0.949 \\
cavitary\_\allowbreak nodule & 263 & 0.010853 & 0.977 & 0.943 & 0.046 & 0.993 \\
cavity & 84 & 0.298880 & 0.976 & 0.994 & 0.124 & 0.998 \\
central\_\allowbreak venous\_\allowbreak catheter & 5,605 & 0.380548 & 0.961 & 0.954 & 0.557 & 0.989 \\
chest\_\allowbreak tube & 4,990 & 0.447080 & 0.967 & 0.967 & 0.609 & 0.993 \\
chest\_\allowbreak wall\_\allowbreak deformity & 1,606 & 0.161256 & 0.933 & 0.940 & 0.208 & 0.982 \\
clavicle\_\allowbreak fracture & 990 & 0.308245 & 0.981 & 0.994 & 0.630 & 0.999 \\
compression\_\allowbreak fracture & 187 & 0.012543 & 0.952 & 0.964 & 0.049 & 0.987 \\
consolidation & 19,660 & 0.483542 & 0.975 & 0.914 & 0.718 & 0.984 \\
copd & 19,941 & 0.469946 & 0.973 & 0.978 & 0.900 & 0.996 \\
degenerative\_\allowbreak changes & 19,647 & 0.645175 & 0.988 & 0.942 & 0.795 & 0.984 \\
demineralization & 165 & 0.006589 & 0.897 & 0.869 & 0.012 & 0.945 \\
dialysis\_\allowbreak catheter & 420 & 0.161744 & 0.990 & 0.988 & 0.263 & 0.999 \\
diaphragmatic\_\allowbreak hernia & 2,574 & 0.319327 & 0.989 & 0.989 & 0.701 & 0.999 \\
diffuse\_\allowbreak pulmonary\_\allowbreak nodules & 501 & 0.094999 & 0.990 & 0.962 & 0.121 & 0.997 \\
elevated\_\allowbreak diaphragm & 6,482 & 0.453617 & 0.989 & 0.995 & 0.925 & 0.999 \\
emphysema & 4,696 & 0.311584 & 0.981 & 0.974 & 0.657 & 0.997 \\
empyema & 101 & 0.025217 & 0.980 & 0.972 & 0.037 & 0.996 \\
end\_\allowbreak of\_\allowbreak the\_\allowbreak vessel & 24 & 0.010803 & 0.917 & 0.826 & 0.001 & 0.923 \\
endotracheal\_\allowbreak tube & 8,335 & 0.489889 & 0.994 & 0.993 & 0.928 & 0.999 \\
endotracheal\_\allowbreak tube\_\allowbreak removed & 296 & 0.034585 & 0.932 & 0.906 & 0.030 & 0.978 \\
enteric\_\allowbreak tube\_\allowbreak removed & 134 & 0.016984 & 0.948 & 0.922 & 0.017 & 0.987 \\
epidural\_\allowbreak catheter & 120 & 0.129113 & 0.992 & 0.998 & 0.391 & 1.000 \\
feeding\_\allowbreak tube & 9,337 & 0.471198 & 0.985 & 0.981 & 0.835 & 0.997 \\
fibro\_\allowbreak lesion & 27 & 0.015748 & 0.889 & 0.870 & 0.002 & 0.937 \\
fibrosis & 11,831 & 0.444876 & 0.985 & 0.978 & 0.853 & 0.997 \\
fissural\_\allowbreak thickening & 26 & 0.127022 & 0.923 & 0.997 & 0.074 & 0.988 \\
granuloma & 5,705 & 0.434366 & 0.995 & 0.996 & 0.943 & 1.000 \\
ground\_\allowbreak glass\_\allowbreak nodule & 185 & 0.076595 & 0.973 & 0.960 & 0.045 & 0.994 \\
ground\_\allowbreak glass\_\allowbreak opacities & 207 & 0.372389 & 1.000 & 0.995 & 0.326 & 0.998 \\
hardware\_\allowbreak wire & 312 & 0.189975 & 0.907 & 0.903 & 0.030 & 0.960 \\
healed\_\allowbreak rib\_\allowbreak fracture & 3,986 & 0.427337 & 0.979 & 0.957 & 0.493 & 0.993 \\
heart\_\allowbreak failure & 994 & 0.219269 & 0.976 & 0.978 & 0.318 & 0.995 \\
hemothorax & 256 & 0.061027 & 0.949 & 0.953 & 0.051 & 0.990 \\
hilar\_\allowbreak enlargement & 2,454 & 0.289998 & 0.943 & 0.954 & 0.345 & 0.989 \\
hilar\_\allowbreak lymphadenopathy & 1,217 & 0.242338 & 0.984 & 0.987 & 0.487 & 0.999 \\
hilar\_\allowbreak mass & 18 & 0.048797 & 1.000 & 0.977 & 0.008 & 0.995 \\
hydropneumothorax & 208 & 0.219123 & 0.990 & 0.994 & 0.251 & 0.999 \\
hyperinflation & 13,114 & 0.578664 & 0.975 & 0.985 & 0.897 & 0.997 \\
infiltrate & 2,855 & 0.439768 & 0.949 & 0.894 & 0.214 & 0.966 \\
interstitial\_\allowbreak lung\_\allowbreak disease & 14,311 & 0.528699 & 0.963 & 0.966 & 0.809 & 0.993 \\
intubation & 41 & 0.021748 & 0.976 & 0.925 & 0.006 & 0.989 \\
kerley\_\allowbreak lines & 189 & 0.086548 & 0.968 & 0.981 & 0.092 & 0.998 \\
kyphosis & 2,202 & 0.577836 & 0.980 & 0.987 & 0.629 & 0.998 \\
lesion & 29 & 0.029299 & 1.000 & 0.914 & 0.004 & 0.991 \\
loop\_\allowbreak recorder & 165 & 0.139448 & 0.982 & 0.999 & 0.545 & 0.997 \\
lordosis\_\allowbreak abnormality & 192 & 0.150946 & 0.943 & 0.971 & 0.062 & 0.988 \\
low\_\allowbreak lung\_\allowbreak volumes & 9,347 & 0.442052 & 0.978 & 0.971 & 0.770 & 0.991 \\
lung\_\allowbreak surgery & 7 & 0.002555 & 1.000 & 0.777 & 0.000 & 0.930 \\
lung\_\allowbreak transplant & 64 & 0.067212 & 0.984 & 0.984 & 0.040 & 0.998 \\
lvad & 55 & 0.040988 & 1.000 & 0.981 & 0.029 & 0.999 \\
lymphadenopathy & 83 & 0.068663 & 1.000 & 0.982 & 0.045 & 0.998 \\
malignancy & 2,034 & 0.206227 & 0.923 & 0.934 & 0.230 & 0.979 \\
mass & 2,465 & 0.317632 & 0.964 & 0.968 & 0.438 & 0.995 \\
mediastinal\_\allowbreak drain & 435 & 0.173036 & 0.998 & 0.979 & 0.176 & 0.999 \\
mediastinal\_\allowbreak mass & 521 & 0.089233 & 0.912 & 0.945 & 0.084 & 0.980 \\
mediastinal\_\allowbreak shift & 546 & 0.171739 & 0.989 & 0.989 & 0.344 & 0.999 \\
mediastinal\_\allowbreak widening & 1,579 & 0.291128 & 0.964 & 0.976 & 0.400 & 0.994 \\
metastasis & 46 & 0.060487 & 1.000 & 0.987 & 0.037 & 0.999 \\
mpa\_\allowbreak dilation & 751 & 0.187853 & 0.963 & 0.973 & 0.219 & 0.994 \\
nipple\_\allowbreak shadows & 245 & 0.118598 & 0.988 & 0.988 & 0.170 & 0.997 \\
opacity & 14,567 & 0.482485 & 0.955 & 0.880 & 0.563 & 0.965 \\
orthopedic\_\allowbreak hardware & 168 & 0.025211 & 0.988 & 0.908 & 0.019 & 0.989 \\
osseous\_\allowbreak structures & 185 & 0.021777 & 0.908 & 0.814 & 0.009 & 0.933 \\
osteopenia & 6,429 & 0.551412 & 0.995 & 0.989 & 0.865 & 0.998 \\
pacemaker & 4,766 & 0.421163 & 0.992 & 0.980 & 0.713 & 0.995 \\
peribronchial\_\allowbreak thickening & 4,120 & 0.427180 & 0.960 & 0.962 & 0.524 & 0.993 \\
pericardial\_\allowbreak effusion & 476 & 0.123608 & 0.960 & 0.974 & 0.155 & 0.994 \\
perihilar\_\allowbreak opacities & 5,610 & 0.332480 & 0.972 & 0.959 & 0.589 & 0.994 \\
picc\_\allowbreak line & 6,458 & 0.399459 & 0.995 & 0.990 & 0.874 & 0.998 \\
pleural\_\allowbreak calcification & 37 & 0.038831 & 1.000 & 0.985 & 0.025 & 0.999 \\
pleural\_\allowbreak effusion & 26,497 & 0.568315 & 0.992 & 0.991 & 0.969 & 0.999 \\
pleural\_\allowbreak plaque & 68 & 0.131458 & 0.985 & 0.992 & 0.078 & 0.998 \\
pleural\_\allowbreak scarring & 93 & 0.019607 & 0.903 & 0.949 & 0.017 & 0.980 \\
pleural\_\allowbreak thickening & 6,398 & 0.403532 & 0.982 & 0.977 & 0.746 & 0.997 \\
pneumomediastinum & 414 & 0.158128 & 0.988 & 0.992 & 0.359 & 0.999 \\
pneumonia & 14,517 & 0.445592 & 0.968 & 0.948 & 0.746 & 0.992 \\
pneumonitis & 301 & 0.084546 & 0.987 & 0.953 & 0.063 & 0.992 \\
pneumoperitoneum & 355 & 0.246684 & 0.977 & 0.993 & 0.340 & 0.997 \\
pneumothorax & 4,189 & 0.411499 & 0.982 & 0.985 & 0.742 & 0.998 \\
port\_\allowbreak catheter & 1,480 & 0.189375 & 0.978 & 0.982 & 0.458 & 0.997 \\
postoperative\_\allowbreak changes & 489 & 0.086523 & 0.957 & 0.948 & 0.087 & 0.989 \\
prosthetic & 485 & 0.306917 & 0.961 & 0.954 & 0.097 & 0.987 \\
prosthetic\_\allowbreak valve & 455 & 0.039775 & 0.958 & 0.902 & 0.045 & 0.983 \\
pulmonary\_\allowbreak artery\_\allowbreak catheter & 567 & 0.243814 & 0.988 & 0.973 & 0.182 & 0.998 \\
pulmonary\_\allowbreak edema & 17,011 & 0.548177 & 0.969 & 0.972 & 0.861 & 0.996 \\
pulmonary\_\allowbreak hypertension & 1,647 & 0.299972 & 0.955 & 0.937 & 0.209 & 0.985 \\
pulmonary\_\allowbreak nodule & 10,154 & 0.407126 & 0.979 & 0.982 & 0.855 & 0.997 \\
pulmonary\_\allowbreak vascular\_\allowbreak congestion & 10,816 & 0.478537 & 0.960 & 0.966 & 0.763 & 0.993 \\
pulmonary\_\allowbreak venous\_\allowbreak hypertension & 528 & 0.109072 & 0.956 & 0.962 & 0.123 & 0.993 \\
reactive\_\allowbreak airways\_\allowbreak disease & 305 & 0.082536 & 0.990 & 0.980 & 0.136 & 0.999 \\
respiratory\_\allowbreak failure & 226 & 0.127586 & 0.965 & 0.961 & 0.056 & 0.992 \\
retrocardiac\_\allowbreak opacity & 881 & 0.158879 & 0.989 & 0.971 & 0.238 & 0.996 \\
rib\_\allowbreak fracture & 5,887 & 0.432311 & 0.990 & 0.979 & 0.753 & 0.998 \\
rotation & 739 & 0.159376 & 0.935 & 0.972 & 0.204 & 0.988 \\
scarring & 1,282 & 0.259624 & 0.995 & 0.947 & 0.205 & 0.978 \\
scoliosis & 5,924 & 0.534002 & 0.991 & 0.998 & 0.959 & 0.999 \\
soft\_\allowbreak tissue\_\allowbreak swelling & 1,558 & 0.344448 & 0.942 & 0.976 & 0.390 & 0.990 \\
solitary\_\allowbreak pulmonary\_\allowbreak nodule & 1,744 & 0.216567 & 0.976 & 0.982 & 0.496 & 0.996 \\
spinal\_\allowbreak hardware & 411 & 0.194528 & 0.961 & 0.967 & 0.112 & 0.990 \\
sternotomy & 3,776 & 0.265287 & 0.979 & 0.976 & 0.623 & 0.995 \\
sternotomy\_\allowbreak wires & 4,306 & 0.361159 & 0.987 & 0.961 & 0.539 & 0.994 \\
subcutaneous\_\allowbreak emphysema & 1,150 & 0.264977 & 0.991 & 0.990 & 0.555 & 0.998 \\
support\_\allowbreak device & 1,764 & 0.253653 & 0.944 & 0.922 & 0.185 & 0.981 \\
surgical\_\allowbreak hardware & 7,087 & 0.375387 & 0.955 & 0.924 & 0.489 & 0.982 \\
tented\_\allowbreak diaphragm & 16 & 0.033279 & 1.000 & 0.989 & 0.016 & 0.998 \\
tracheal\_\allowbreak shift & 281 & 0.103954 & 0.975 & 0.978 & 0.119 & 0.996 \\
tracheostomy & 290 & 0.044327 & 0.972 & 0.961 & 0.071 & 0.993 \\
tracheostomy\_\allowbreak tube & 1,567 & 0.286325 & 0.996 & 0.993 & 0.705 & 1.000 \\
\end{longtable}
\endgroup
\clearpage
\begingroup
\small\setlength{\tabcolsep}{2pt}\renewcommand{\arraystretch}{1.10}
\begin{longtable}{@{}>{\raggedright\arraybackslash}p{5.4cm}rrrrrr@{}}
\caption{CT per-tag results (256 tags) at test-selected maximum-Youden thresholds. Support is reference-positive count; Sens. and Spec. denote sensitivity and specificity.}\label{tab:ct-per-tag}\\
\toprule
Tag & Support & Threshold & Sens. & Spec. & F1 & AUROC \\
\midrule
\endfirsthead
\multicolumn{7}{l}{\textit{CT: test-selected maximum-Youden thresholds (continued)}}\\
\toprule
Tag & Support & Threshold & Sens. & Spec. & F1 & AUROC \\
\midrule
\endhead
\midrule
\multicolumn{7}{r}{\textit{Continued on next page}}\\
\endfoot
\bottomrule
\endlastfoot
ABDOMINAL\_\allowbreak WALL\_\allowbreak HERNIA & 72 & 0.128230 & 0.958 & 0.971 & 0.142 & 0.994 \\
ABERRANT\_\allowbreak RIGHT\_\allowbreak SUBCLAVIAN\_\allowbreak ARTERY & 161 & 0.112896 & 0.944 & 0.967 & 0.241 & 0.990 \\
ABSENT\_\allowbreak GALLBLADDER & 41 & 0.028526 & 0.780 & 0.879 & 0.018 & 0.909 \\
ADRENAL\_\allowbreak ADENOMA & 429 & 0.389781 & 0.970 & 0.987 & 0.692 & 0.997 \\
ADRENAL\_\allowbreak CALCIFICATION & 25 & 0.029913 & 0.880 & 0.953 & 0.032 & 0.964 \\
ADRENAL\_\allowbreak ENLARGEMENT & 11 & 0.035492 & 1.000 & 0.971 & 0.026 & 0.994 \\
ADRENAL\_\allowbreak LESION & 16 & 0.013552 & 0.938 & 0.928 & 0.014 & 0.973 \\
ADRENAL\_\allowbreak MASS & 46 & 0.083183 & 0.935 & 0.965 & 0.079 & 0.969 \\
ADRENAL\_\allowbreak MYELOLIPOMA & 22 & 0.280893 & 1.000 & 0.998 & 0.506 & 1.000 \\
ADRENAL\_\allowbreak NODULE & 619 & 0.281837 & 0.947 & 0.946 & 0.430 & 0.984 \\
ADRENAL\_\allowbreak THICKENING & 415 & 0.218875 & 0.959 & 0.936 & 0.303 & 0.984 \\
AIR\_\allowbreak BRONCHOGRAM & 416 & 0.243645 & 0.913 & 0.906 & 0.220 & 0.971 \\
AIR\_\allowbreak TRAPPING & 327 & 0.245892 & 0.957 & 0.970 & 0.419 & 0.993 \\
AIRSPACE\_\allowbreak OPACITY & 1,080 & 0.357337 & 0.914 & 0.897 & 0.402 & 0.966 \\
AIRWAY\_\allowbreak NARROWING & 380 & 0.196368 & 0.934 & 0.923 & 0.243 & 0.974 \\
ANASARCA & 32 & 0.042680 & 0.969 & 0.963 & 0.056 & 0.995 \\
ANATOMIC\_\allowbreak VARIATION & 29 & 0.125637 & 0.966 & 0.983 & 0.101 & 0.987 \\
AORTIC\_\allowbreak ANEURYSM & 2,599 & 0.432793 & 0.979 & 0.954 & 0.802 & 0.993 \\
AORTIC\_\allowbreak DILATATION & 68 & 0.021542 & 0.912 & 0.823 & 0.024 & 0.938 \\
AORTIC\_\allowbreak DISSECTION & 148 & 0.269136 & 0.980 & 0.994 & 0.639 & 0.996 \\
AORTIC\_\allowbreak ECTASIA & 71 & 0.039167 & 0.986 & 0.918 & 0.056 & 0.982 \\
AORTIC\_\allowbreak FILLING\_\allowbreak DEFECT & 32 & 0.034890 & 0.969 & 0.936 & 0.033 & 0.982 \\
AORTIC\_\allowbreak TORTUOSITY & 57 & 0.038682 & 0.947 & 0.949 & 0.068 & 0.984 \\
AORTIC\_\allowbreak VALVE\_\allowbreak CALCIFICATION & 547 & 0.265471 & 0.969 & 0.957 & 0.466 & 0.986 \\
ARCHITECTURAL\_\allowbreak DISTORTION & 350 & 0.139904 & 0.883 & 0.858 & 0.132 & 0.946 \\
ASCITES & 148 & 0.322090 & 0.980 & 0.993 & 0.588 & 0.997 \\
ATELECTASIS & 8,917 & 0.554926 & 0.975 & 0.972 & 0.957 & 0.994 \\
ATHEROSCLEROSIS & 10,360 & 0.601265 & 0.886 & 0.894 & 0.855 & 0.958 \\
AZYGOS\_\allowbreak LOBE & 61 & 0.058847 & 0.967 & 0.960 & 0.094 & 0.990 \\
BILIARY\_\allowbreak DILATATION & 9 & 0.005955 & 1.000 & 0.837 & 0.004 & 0.972 \\
BLADDER\_\allowbreak WALL\_\allowbreak THICKENING & 63 & 0.032518 & 0.984 & 0.929 & 0.058 & 0.991 \\
BLEB & 101 & 0.090644 & 0.941 & 0.972 & 0.191 & 0.991 \\
BOCHDALEK\_\allowbreak HERNIA & 68 & 0.395664 & 1.000 & 0.999 & 0.889 & 1.000 \\
BONE\_\allowbreak METASTASIS & 385 & 0.236614 & 0.956 & 0.969 & 0.455 & 0.992 \\
BREAST\_\allowbreak ASYMMETRY & 19 & 0.021774 & 0.789 & 0.943 & 0.018 & 0.925 \\
BREAST\_\allowbreak CALCIFICATION & 82 & 0.088956 & 0.951 & 0.944 & 0.089 & 0.978 \\
BREAST\_\allowbreak IMPLANT & 387 & 0.284957 & 0.917 & 0.980 & 0.549 & 0.982 \\
BREAST\_\allowbreak MASS & 17 & 0.020038 & 1.000 & 0.971 & 0.039 & 0.996 \\
BREAST\_\allowbreak NODULARITY & 15 & 0.016777 & 0.867 & 0.921 & 0.011 & 0.952 \\
BREAST\_\allowbreak NODULE & 100 & 0.091493 & 0.970 & 0.958 & 0.139 & 0.993 \\
BRONCHIAL\_\allowbreak DIVERTICULUM & 79 & 0.136923 & 0.962 & 0.986 & 0.270 & 0.996 \\
BRONCHIAL\_\allowbreak WALL\_\allowbreak THICKENING & 4,537 & 0.513981 & 0.982 & 0.973 & 0.925 & 0.997 \\
BRONCHIECTASIS & 1,917 & 0.487330 & 0.984 & 0.989 & 0.920 & 0.998 \\
BRONCHIOLITIS & 1,250 & 0.354759 & 0.938 & 0.916 & 0.495 & 0.977 \\
BULLA & 19 & 0.050535 & 0.947 & 0.975 & 0.048 & 0.945 \\
CALCIFICATION & 452 & 0.135968 & 0.918 & 0.729 & 0.098 & 0.893 \\
CALCIFIED\_\allowbreak GRANULOMA & 3,794 & 0.591008 & 0.957 & 0.975 & 0.903 & 0.993 \\
CALCIFIED\_\allowbreak LYMPH\_\allowbreak NODE & 270 & 0.161335 & 0.911 & 0.913 & 0.165 & 0.967 \\
CALCIFIED\_\allowbreak NODULE & 1,904 & 0.388237 & 0.884 & 0.828 & 0.412 & 0.928 \\
CARDIAC\_\allowbreak CHAMBER\_\allowbreak ENLARGEMENT & 357 & 0.208194 & 0.941 & 0.925 & 0.238 & 0.985 \\
CARDIAC\_\allowbreak DEVICE & 76 & 0.058399 & 0.868 & 0.908 & 0.047 & 0.951 \\
CARDIOMEGALY & 2,365 & 0.519739 & 0.991 & 0.996 & 0.972 & 0.999 \\
CAVITARY\_\allowbreak NODULE & 253 & 0.161050 & 0.901 & 0.898 & 0.135 & 0.965 \\
CAVITY & 378 & 0.217319 & 0.876 & 0.918 & 0.218 & 0.955 \\
CENTRAL\_\allowbreak VENOUS\_\allowbreak CATHETER & 36 & 0.037040 & 0.917 & 0.930 & 0.032 & 0.948 \\
CERVICAL\_\allowbreak RIB & 16 & 0.036825 & 0.938 & 0.974 & 0.038 & 0.988 \\
CHEST\_\allowbreak PORT & 168 & 0.075856 & 0.976 & 0.929 & 0.139 & 0.987 \\
CHOLECYSTECTOMY & 679 & 0.216978 & 0.825 & 0.850 & 0.206 & 0.903 \\
CHOLECYSTECTOMY\_\allowbreak CLIP & 294 & 0.189971 & 0.925 & 0.937 & 0.232 & 0.974 \\
CHOLELITHIASIS & 1,384 & 0.359886 & 0.987 & 0.971 & 0.769 & 0.997 \\
CHRONIC\_\allowbreak BRONCHIAL\_\allowbreak INFLAMMATORY\_\allowbreak CHANGE & 11 & 0.009851 & 1.000 & 0.927 & 0.010 & 0.988 \\
CHRONIC\_\allowbreak PANCREATITIS & 16 & 0.114717 & 1.000 & 0.995 & 0.182 & 0.999 \\
CHRONIC\_\allowbreak PULMONARY\_\allowbreak EMBOLISM & 59 & 0.024366 & 1.000 & 0.907 & 0.042 & 0.992 \\
CIRRHOSIS & 187 & 0.189388 & 0.963 & 0.981 & 0.395 & 0.991 \\
CLAVICLE\_\allowbreak FRACTURE & 25 & 0.029371 & 1.000 & 0.960 & 0.042 & 0.996 \\
COLONIC\_\allowbreak DIVERTICULOSIS & 697 & 0.248559 & 0.974 & 0.949 & 0.486 & 0.990 \\
COLONIC\_\allowbreak INTERPOSITION & 21 & 0.020783 & 0.857 & 0.918 & 0.015 & 0.954 \\
COMPLETE\_\allowbreak RESPONSE & 79 & 0.031411 & 0.899 & 0.801 & 0.024 & 0.931 \\
CONGESTIVE\_\allowbreak HEART\_\allowbreak FAILURE & 104 & 0.092975 & 0.981 & 0.969 & 0.189 & 0.995 \\
CONSOLIDATION & 1,480 & 0.412650 & 0.957 & 0.968 & 0.751 & 0.994 \\
COPD & 2,801 & 0.413492 & 0.974 & 0.791 & 0.499 & 0.935 \\
CORONARY\_\allowbreak ARTERY\_\allowbreak BYPASS\_\allowbreak GRAFT & 69 & 0.030799 & 0.942 & 0.850 & 0.029 & 0.952 \\
CORONARY\_\allowbreak ARTERY\_\allowbreak CALCIFICATION & 15,950 & 0.625553 & 0.961 & 0.946 & 0.959 & 0.990 \\
CORONARY\_\allowbreak ARTERY\_\allowbreak DISEASE & 57 & 0.034967 & 0.877 & 0.893 & 0.031 & 0.929 \\
CORONARY\_\allowbreak ARTERY\_\allowbreak STENT & 249 & 0.198687 & 0.940 & 0.961 & 0.293 & 0.984 \\
DEGENERATIVE\_\allowbreak SPINE\_\allowbreak CHANGES & 7,748 & 0.563767 & 0.969 & 0.973 & 0.949 & 0.995 \\
DIAPHRAGM\_\allowbreak ELEVATION & 421 & 0.318923 & 0.960 & 0.988 & 0.700 & 0.993 \\
DIAPHRAGMATIC\_\allowbreak EVENTRATION & 102 & 0.117843 & 0.971 & 0.977 & 0.234 & 0.996 \\
DIAPHRAGMATIC\_\allowbreak EVENTRATION\_\allowbreak OR\_\allowbreak HERNIA & 12 & 0.018944 & 1.000 & 0.957 & 0.019 & 0.994 \\
DIAPHRAGMATIC\_\allowbreak HERNIA & 102 & 0.120817 & 0.912 & 0.966 & 0.160 & 0.982 \\
DISTANT\_\allowbreak METASTASIS & 945 & 0.315320 & 0.940 & 0.921 & 0.442 & 0.981 \\
DUODENAL\_\allowbreak DIVERTICULUM & 68 & 0.103702 & 0.985 & 0.984 & 0.229 & 0.998 \\
EMPHYSEMA & 8,293 & 0.692679 & 0.990 & 0.996 & 0.990 & 0.999 \\
ENDOBRONCHIAL\_\allowbreak LESION & 266 & 0.150921 & 0.951 & 0.908 & 0.162 & 0.978 \\
ENDOTRACHEAL\_\allowbreak TUBE & 19 & 0.041302 & 0.947 & 0.966 & 0.035 & 0.991 \\
ESOPHAGEAL\_\allowbreak DILATION & 349 & 0.205655 & 0.977 & 0.969 & 0.438 & 0.995 \\
ESOPHAGEAL\_\allowbreak MASS & 82 & 0.066406 & 0.976 & 0.955 & 0.111 & 0.991 \\
ESOPHAGEAL\_\allowbreak WALL\_\allowbreak THICKENING & 173 & 0.170910 & 0.994 & 0.978 & 0.355 & 0.998 \\
FIBROSIS & 2,361 & 0.416715 & 0.927 & 0.914 & 0.644 & 0.976 \\
GALLBLADDER\_\allowbreak SLUDGE & 20 & 0.073455 & 0.950 & 0.988 & 0.102 & 0.994 \\
GALLBLADDER\_\allowbreak WALL\_\allowbreak THICKENING & 28 & 0.059564 & 0.893 & 0.983 & 0.093 & 0.982 \\
GASTRIC\_\allowbreak DIVERTICULUM & 35 & 0.115027 & 0.943 & 0.991 & 0.207 & 0.989 \\
GRANULOMA & 1,760 & 0.319308 & 0.966 & 0.877 & 0.502 & 0.960 \\
GROUND\_\allowbreak GLASS\_\allowbreak NODULE & 2,178 & 0.426884 & 0.945 & 0.914 & 0.633 & 0.978 \\
GROUND\_\allowbreak GLASS\_\allowbreak OPACITY & 4,000 & 0.544718 & 0.948 & 0.958 & 0.860 & 0.988 \\
GYNECOMASTIA & 593 & 0.439056 & 0.987 & 0.998 & 0.948 & 0.998 \\
HEMOTHORAX & 53 & 0.080775 & 0.981 & 0.986 & 0.209 & 0.998 \\
HEPATIC\_\allowbreak CALCIFICATION & 77 & 0.068209 & 0.857 & 0.882 & 0.038 & 0.927 \\
HEPATIC\_\allowbreak CALCIFIED\_\allowbreak GRANULOMA & 86 & 0.108299 & 0.919 & 0.943 & 0.088 & 0.977 \\
HEPATIC\_\allowbreak CYST & 1,813 & 0.471856 & 0.955 & 0.961 & 0.754 & 0.990 \\
HEPATIC\_\allowbreak HEMANGIOMA & 100 & 0.068924 & 1.000 & 0.949 & 0.122 & 0.995 \\
HEPATIC\_\allowbreak HYPODENSITY & 260 & 0.143886 & 0.896 & 0.900 & 0.140 & 0.958 \\
HEPATIC\_\allowbreak LESION & 320 & 0.171728 & 0.922 & 0.914 & 0.194 & 0.968 \\
HEPATIC\_\allowbreak MASS & 20 & 0.034153 & 0.950 & 0.970 & 0.042 & 0.953 \\
HEPATIC\_\allowbreak NODULE & 10 & 0.007328 & 1.000 & 0.849 & 0.005 & 0.976 \\
HEPATIC\_\allowbreak STEATOSIS & 2,159 & 0.403479 & 0.985 & 0.973 & 0.852 & 0.996 \\
HEPATOMEGALY & 100 & 0.064272 & 0.960 & 0.921 & 0.078 & 0.989 \\
HEPATOSPLENOMEGALY & 17 & 0.022690 & 1.000 & 0.959 & 0.028 & 0.994 \\
HIATAL\_\allowbreak HERNIA & 3,042 & 0.554350 & 0.983 & 0.993 & 0.964 & 0.998 \\
HONEYCOMBING & 337 & 0.266672 & 0.970 & 0.987 & 0.641 & 0.997 \\
HYDRONEPHROSIS & 30 & 0.035974 & 0.900 & 0.974 & 0.067 & 0.979 \\
HYDROPNEUMOTHORAX & 46 & 0.064124 & 1.000 & 0.974 & 0.110 & 0.999 \\
HYPERINFLATION & 283 & 0.202240 & 0.961 & 0.970 & 0.386 & 0.991 \\
HYSTERECTOMY & 13 & 0.002741 & 0.923 & 0.762 & 0.004 & 0.889 \\
INFARCTED\_\allowbreak LUNG & 48 & 0.094821 & 0.958 & 0.992 & 0.279 & 0.994 \\
INFECTION & 538 & 0.237346 & 0.922 & 0.885 & 0.232 & 0.960 \\
INGUINAL\_\allowbreak HERNIA & 46 & 0.020582 & 1.000 & 0.890 & 0.028 & 0.987 \\
INTERSTITIAL\_\allowbreak EDEMA & 145 & 0.194734 & 0.979 & 0.983 & 0.371 & 0.997 \\
INTERSTITIAL\_\allowbreak LUNG\_\allowbreak DISEASE & 2,047 & 0.446443 & 0.940 & 0.942 & 0.697 & 0.984 \\
INTERVENTRICULAR\_\allowbreak SEPTAL\_\allowbreak BOWING & 67 & 0.472020 & 0.970 & 0.999 & 0.793 & 0.994 \\
INTRAPULMONARY\_\allowbreak LYMPH\_\allowbreak NODE & 28 & 0.017235 & 0.857 & 0.885 & 0.014 & 0.939 \\
IVC\_\allowbreak FILTER & 31 & 0.057605 & 0.871 & 0.971 & 0.061 & 0.968 \\
LEFT\_\allowbreak ATRIAL\_\allowbreak ENLARGEMENT & 130 & 0.184287 & 0.954 & 0.976 & 0.263 & 0.994 \\
LEFT\_\allowbreak VENTRICULAR\_\allowbreak ENLARGEMENT & 42 & 0.054142 & 0.976 & 0.933 & 0.041 & 0.988 \\
LIPOMA & 106 & 0.137488 & 0.991 & 0.976 & 0.238 & 0.998 \\
LOBAR\_\allowbreak COLLAPSE & 140 & 0.130229 & 0.936 & 0.955 & 0.168 & 0.982 \\
LOOP\_\allowbreak RECORDER & 195 & 0.195966 & 0.979 & 0.979 & 0.386 & 0.995 \\
LUNG\_\allowbreak MASS & 1,070 & 0.345660 & 0.947 & 0.933 & 0.517 & 0.983 \\
LUNG\_\allowbreak RADS & 103 & 0.063429 & 0.845 & 0.855 & 0.040 & 0.909 \\
LYMPHADENOPATHY & 3,566 & 0.466971 & 0.966 & 0.932 & 0.791 & 0.986 \\
LYMPHANGITIC\_\allowbreak CARCINOMATOSIS & 37 & 0.088580 & 0.919 & 0.973 & 0.081 & 0.989 \\
LYTIC\_\allowbreak LESION & 307 & 0.223189 & 0.893 & 0.953 & 0.285 & 0.972 \\
MASTECTOMY & 61 & 0.062121 & 0.820 & 0.955 & 0.071 & 0.946 \\
MEDIASTINAL\_\allowbreak MASS & 476 & 0.259934 & 0.861 & 0.923 & 0.269 & 0.957 \\
MEDIASTINAL\_\allowbreak SHIFT & 34 & 0.146247 & 0.941 & 0.992 & 0.212 & 0.994 \\
METASTATIC\_\allowbreak NODULES & 423 & 0.279010 & 0.931 & 0.929 & 0.280 & 0.976 \\
MITRAL\_\allowbreak ANNULAR\_\allowbreak CALCIFICATION & 510 & 0.276782 & 0.976 & 0.973 & 0.566 & 0.994 \\
MOSAIC\_\allowbreak ATTENUATION & 484 & 0.330429 & 0.975 & 0.985 & 0.682 & 0.997 \\
MUCOUS\_\allowbreak PLUGGING & 2,083 & 0.462290 & 0.959 & 0.969 & 0.813 & 0.992 \\
MULTIPLE\_\allowbreak PULMONARY\_\allowbreak NODULES & 11,887 & 0.632974 & 0.914 & 0.938 & 0.913 & 0.978 \\
NEPHRECTOMY & 37 & 0.076926 & 0.703 & 0.964 & 0.048 & 0.890 \\
NEPHROLITHIASIS & 790 & 0.341112 & 0.948 & 0.967 & 0.608 & 0.990 \\
NODAL\_\allowbreak METASTASIS & 742 & 0.292070 & 0.968 & 0.945 & 0.479 & 0.990 \\
NONSPECIFIC\_\allowbreak INTERSTITIAL\_\allowbreak PNEUMONIA\_\allowbreak PATTERN & 28 & 0.031931 & 0.964 & 0.934 & 0.028 & 0.991 \\
NORMAL & 513 & 0.281481 & 0.975 & 0.979 & 0.629 & 0.997 \\
OPACITY & 2,116 & 0.385905 & 0.869 & 0.776 & 0.371 & 0.898 \\
ORGANIZING\_\allowbreak PNEUMONIA\_\allowbreak PATTERN & 26 & 0.022345 & 0.885 & 0.921 & 0.020 & 0.970 \\
OSSEOUS\_\allowbreak DEMINERALIZATION & 164 & 0.223638 & 0.963 & 0.980 & 0.352 & 0.995 \\
OSTEOPENIA & 387 & 0.283023 & 0.972 & 0.982 & 0.592 & 0.996 \\
OSTEOPOROSIS & 19 & 0.049329 & 0.947 & 0.983 & 0.070 & 0.987 \\
PACEMAKER & 406 & 0.195120 & 0.966 & 0.970 & 0.479 & 0.992 \\
PANCREATIC\_\allowbreak ATROPHY & 91 & 0.060399 & 0.989 & 0.936 & 0.090 & 0.991 \\
PANCREATIC\_\allowbreak CALCIFICATION & 70 & 0.091220 & 0.900 & 0.955 & 0.089 & 0.972 \\
PANCREATIC\_\allowbreak CYST & 66 & 0.072226 & 0.909 & 0.963 & 0.102 & 0.973 \\
PANCREATIC\_\allowbreak STEATOSIS & 45 & 0.108788 & 0.956 & 0.979 & 0.124 & 0.993 \\
PARTIAL\_\allowbreak RESPONSE & 201 & 0.134204 & 0.891 & 0.913 & 0.125 & 0.958 \\
PECTUS\_\allowbreak EXCAVATUM & 77 & 0.388307 & 1.000 & 0.999 & 0.895 & 1.000 \\
PERICARDIAL\_\allowbreak CALCIFICATION & 35 & 0.060712 & 0.943 & 0.957 & 0.051 & 0.980 \\
PERICARDIAL\_\allowbreak CYST & 16 & 0.022402 & 0.875 & 0.951 & 0.020 & 0.921 \\
PERICARDIAL\_\allowbreak EFFUSION & 1,272 & 0.457011 & 0.986 & 0.990 & 0.894 & 0.997 \\
PERICARDIAL\_\allowbreak THICKENING & 45 & 0.109524 & 0.978 & 0.983 & 0.150 & 0.996 \\
PERINEPHRIC\_\allowbreak STRANDING & 11 & 0.107698 & 0.818 & 0.990 & 0.058 & 0.965 \\
PLEURAL\_\allowbreak BASED\_\allowbreak NODULE & 2,896 & 0.465108 & 0.864 & 0.811 & 0.487 & 0.919 \\
PLEURAL\_\allowbreak CALCIFICATION & 475 & 0.242310 & 0.914 & 0.929 & 0.300 & 0.976 \\
PLEURAL\_\allowbreak EFFUSION & 2,205 & 0.498211 & 0.993 & 0.995 & 0.968 & 0.999 \\
PLEURAL\_\allowbreak METASTASIS & 58 & 0.074318 & 0.966 & 0.949 & 0.071 & 0.987 \\
PLEURAL\_\allowbreak NODULE & 577 & 0.241985 & 0.854 & 0.786 & 0.140 & 0.898 \\
PLEURAL\_\allowbreak PLAQUE & 327 & 0.302224 & 0.957 & 0.988 & 0.643 & 0.996 \\
PLEURAL\_\allowbreak THICKENING & 2,162 & 0.451508 & 0.927 & 0.947 & 0.719 & 0.983 \\
PNEUMOBILIA & 104 & 0.094885 & 0.990 & 0.971 & 0.198 & 0.998 \\
PNEUMOMEDIASTINUM & 78 & 0.243017 & 0.987 & 0.998 & 0.733 & 0.997 \\
PNEUMONIA & 672 & 0.259049 & 0.961 & 0.924 & 0.374 & 0.984 \\
PNEUMOTHORAX & 186 & 0.134708 & 0.984 & 0.981 & 0.404 & 0.998 \\
PORTAL\_\allowbreak HYPERTENSION & 20 & 0.031863 & 1.000 & 0.975 & 0.053 & 0.998 \\
POSTSURGICAL\_\allowbreak CHANGE & 386 & 0.141782 & 0.834 & 0.781 & 0.094 & 0.879 \\
PRIMARY\_\allowbreak TUMOR & 1,551 & 0.429499 & 0.886 & 0.918 & 0.535 & 0.961 \\
PROGRESSIVE\_\allowbreak DISEASE & 872 & 0.304374 & 0.936 & 0.884 & 0.332 & 0.966 \\
PROSTATOMEGALY & 88 & 0.105406 & 0.955 & 0.970 & 0.163 & 0.992 \\
PROSTHETIC\_\allowbreak AORTIC\_\allowbreak VALVE & 106 & 0.108812 & 0.934 & 0.965 & 0.166 & 0.985 \\
PULMONARY\_\allowbreak ARTERY\_\allowbreak ENLARGEMENT & 1,134 & 0.391352 & 0.975 & 0.977 & 0.768 & 0.996 \\
PULMONARY\_\allowbreak ARTERY\_\allowbreak FILLING\_\allowbreak DEFECT & 302 & 0.248467 & 0.974 & 0.985 & 0.572 & 0.995 \\
PULMONARY\_\allowbreak CYST & 87 & 0.079671 & 0.931 & 0.916 & 0.063 & 0.972 \\
PULMONARY\_\allowbreak EDEMA & 128 & 0.127337 & 1.000 & 0.971 & 0.239 & 0.997 \\
PULMONARY\_\allowbreak EMBOLISM & 407 & 0.396060 & 0.985 & 0.998 & 0.938 & 0.999 \\
PULMONARY\_\allowbreak HYPERTENSION & 671 & 0.406026 & 0.981 & 0.983 & 0.733 & 0.995 \\
PULMONARY\_\allowbreak NODULE & 9,358 & 0.586001 & 0.849 & 0.746 & 0.716 & 0.886 \\
RENAL\_\allowbreak ANGIOMYOLIPOMA & 18 & 0.011078 & 0.944 & 0.910 & 0.013 & 0.981 \\
RENAL\_\allowbreak ARTERY\_\allowbreak ANEURYSM & 13 & 0.017908 & 1.000 & 0.964 & 0.025 & 0.992 \\
RENAL\_\allowbreak ATROPHY & 80 & 0.038800 & 0.963 & 0.908 & 0.055 & 0.976 \\
RENAL\_\allowbreak CORTICAL\_\allowbreak SCARRING & 15 & 0.015863 & 0.933 & 0.924 & 0.013 & 0.966 \\
RENAL\_\allowbreak CYST & 2,590 & 0.483311 & 0.946 & 0.934 & 0.726 & 0.981 \\
RENAL\_\allowbreak LESION & 208 & 0.120943 & 0.976 & 0.912 & 0.139 & 0.983 \\
RENAL\_\allowbreak MASS & 17 & 0.055994 & 1.000 & 0.977 & 0.049 & 0.996 \\
RETICULATION & 1,566 & 0.378412 & 0.965 & 0.951 & 0.686 & 0.991 \\
RETROAREOLAR\_\allowbreak BREAST\_\allowbreak TISSUE & 16 & 0.027880 & 0.938 & 0.947 & 0.019 & 0.970 \\
RIB\_\allowbreak DEFORMITY & 18 & 0.003786 & 1.000 & 0.804 & 0.006 & 0.960 \\
RIB\_\allowbreak FRACTURE & 1,250 & 0.403496 & 0.966 & 0.982 & 0.816 & 0.996 \\
RIGHT\_\allowbreak HEART\_\allowbreak ENLARGEMENT & 76 & 0.052224 & 0.987 & 0.938 & 0.079 & 0.990 \\
RIGHT\_\allowbreak VENTRICULAR\_\allowbreak ENLARGEMENT & 108 & 0.106677 & 0.972 & 0.965 & 0.175 & 0.993 \\
SCARRING & 9,001 & 0.559521 & 0.936 & 0.920 & 0.887 & 0.977 \\
SCHMORLS\_\allowbreak NODE & 202 & 0.202804 & 0.946 & 0.974 & 0.335 & 0.986 \\
SCLEROTIC\_\allowbreak LESION & 624 & 0.366073 & 0.925 & 0.970 & 0.563 & 0.986 \\
SCOLIOSIS & 353 & 0.146344 & 0.949 & 0.955 & 0.341 & 0.986 \\
SEPTAL\_\allowbreak THICKENING & 568 & 0.302283 & 0.961 & 0.958 & 0.477 & 0.989 \\
SHOULDER\_\allowbreak ARTHROPLASTY & 51 & 0.016959 & 0.941 & 0.859 & 0.023 & 0.962 \\
SHOULDER\_\allowbreak DEGENERATIVE\_\allowbreak CHANGE & 142 & 0.124121 & 0.972 & 0.957 & 0.184 & 0.990 \\
SIGMOID\_\allowbreak DIVERTICULOSIS & 43 & 0.056613 & 1.000 & 0.975 & 0.106 & 0.996 \\
SLEEVE\_\allowbreak GASTRECTOMY & 21 & 0.022621 & 0.905 & 0.962 & 0.034 & 0.980 \\
SOLID\_\allowbreak NODULE & 4,154 & 0.500625 & 0.850 & 0.723 & 0.488 & 0.873 \\
SPINAL\_\allowbreak HARDWARE & 32 & 0.014069 & 0.812 & 0.829 & 0.010 & 0.895 \\
SPINAL\_\allowbreak STIMULATOR & 17 & 0.038347 & 1.000 & 0.983 & 0.066 & 0.996 \\
SPLENECTOMY & 14 & 0.006382 & 0.929 & 0.775 & 0.004 & 0.916 \\
SPLENIC\_\allowbreak ARTERY\_\allowbreak ANEURYSM & 96 & 0.140794 & 0.979 & 0.982 & 0.267 & 0.997 \\
SPLENIC\_\allowbreak CALCIFICATION & 76 & 0.039398 & 0.921 & 0.883 & 0.040 & 0.962 \\
SPLENIC\_\allowbreak CYST & 68 & 0.066370 & 0.941 & 0.952 & 0.086 & 0.984 \\
SPLENIC\_\allowbreak GRANULOMA & 175 & 0.111098 & 0.971 & 0.928 & 0.142 & 0.987 \\
SPLENIC\_\allowbreak LESION & 40 & 0.095042 & 0.950 & 0.978 & 0.108 & 0.969 \\
SPLENOMEGALY & 334 & 0.206795 & 0.964 & 0.973 & 0.456 & 0.995 \\
SPLENULE & 83 & 0.087047 & 0.940 & 0.934 & 0.076 & 0.971 \\
STABLE\_\allowbreak DISEASE & 2,686 & 0.457842 & 0.831 & 0.803 & 0.445 & 0.889 \\
STERNAL\_\allowbreak FRACTURE & 84 & 0.113545 & 0.976 & 0.979 & 0.218 & 0.995 \\
STERNOTOMY & 203 & 0.075205 & 0.931 & 0.876 & 0.097 & 0.961 \\
STERNOTOMY\_\allowbreak WIRE & 126 & 0.124842 & 0.849 & 0.954 & 0.139 & 0.957 \\
SUBCUTANEOUS\_\allowbreak CYST & 85 & 0.166854 & 0.906 & 0.987 & 0.290 & 0.988 \\
SUBCUTANEOUS\_\allowbreak EDEMA & 14 & 0.062049 & 1.000 & 0.988 & 0.074 & 0.998 \\
SUBCUTANEOUS\_\allowbreak EMPHYSEMA & 58 & 0.093541 & 0.983 & 0.985 & 0.205 & 0.998 \\
SUBCUTANEOUS\_\allowbreak NODULE & 44 & 0.040100 & 0.886 & 0.925 & 0.035 & 0.965 \\
SUBSOLID\_\allowbreak NODULE & 810 & 0.306045 & 0.923 & 0.865 & 0.282 & 0.962 \\
SURGICAL\_\allowbreak CLIP & 134 & 0.042417 & 0.813 & 0.827 & 0.042 & 0.891 \\
SURGICAL\_\allowbreak HARDWARE & 28 & 0.013596 & 0.964 & 0.876 & 0.015 & 0.972 \\
SUSPECTED\_\allowbreak PRIMARY\_\allowbreak LUNG\_\allowbreak CANCER & 1,655 & 0.436548 & 0.921 & 0.921 & 0.575 & 0.975 \\
THYMIC\_\allowbreak HYPERPLASIA & 125 & 0.089901 & 0.936 & 0.961 & 0.173 & 0.982 \\
THYMIC\_\allowbreak MASS & 67 & 0.097408 & 0.955 & 0.981 & 0.190 & 0.983 \\
THYROID\_\allowbreak ABNORMALITY & 381 & 0.187692 & 0.887 & 0.888 & 0.174 & 0.945 \\
THYROID\_\allowbreak ASYMMETRY & 21 & 0.011222 & 0.857 & 0.888 & 0.011 & 0.887 \\
THYROID\_\allowbreak ATROPHY & 98 & 0.064413 & 0.959 & 0.917 & 0.074 & 0.980 \\
THYROID\_\allowbreak CALCIFICATION & 122 & 0.132792 & 0.902 & 0.938 & 0.110 & 0.971 \\
THYROID\_\allowbreak ENLARGEMENT & 226 & 0.088064 & 0.965 & 0.914 & 0.152 & 0.982 \\
THYROID\_\allowbreak HETEROGENEITY & 372 & 0.228191 & 0.968 & 0.948 & 0.329 & 0.987 \\
THYROID\_\allowbreak HYPODENSITY & 59 & 0.051453 & 0.966 & 0.922 & 0.049 & 0.972 \\
THYROID\_\allowbreak LESION & 47 & 0.011775 & 1.000 & 0.859 & 0.023 & 0.973 \\
THYROID\_\allowbreak MASS & 20 & 0.044037 & 0.950 & 0.955 & 0.029 & 0.985 \\
THYROID\_\allowbreak NODULE & 1,763 & 0.425745 & 0.961 & 0.958 & 0.738 & 0.989 \\
THYROIDECTOMY & 27 & 0.047059 & 0.593 & 0.933 & 0.016 & 0.839 \\
TRACHEAL\_\allowbreak ABNORMALITY & 617 & 0.274379 & 0.904 & 0.923 & 0.335 & 0.971 \\
TREE\_\allowbreak IN\_\allowbreak BUD\_\allowbreak OPACITY & 927 & 0.395316 & 0.962 & 0.977 & 0.729 & 0.994 \\
UMBILICAL\_\allowbreak HERNIA & 78 & 0.127707 & 0.962 & 0.973 & 0.161 & 0.993 \\
USUAL\_\allowbreak INTERSTITIAL\_\allowbreak PNEUMONIA\_\allowbreak PATTERN & 188 & 0.166740 & 0.963 & 0.969 & 0.292 & 0.995 \\
UTERINE\_\allowbreak FIBROID & 15 & 0.057312 & 0.867 & 0.979 & 0.042 & 0.971 \\
VALVULAR\_\allowbreak CALCIFICATION & 74 & 0.082027 & 0.905 & 0.943 & 0.076 & 0.976 \\
VASCULAR\_\allowbreak CALCIFICATION & 8,179 & 0.599782 & 0.852 & 0.862 & 0.775 & 0.936 \\
VASCULAR\_\allowbreak CONGESTION & 79 & 0.116674 & 0.962 & 0.978 & 0.193 & 0.991 \\
VASCULAR\_\allowbreak OCCLUSION & 131 & 0.081961 & 0.893 & 0.902 & 0.077 & 0.956 \\
VASCULAR\_\allowbreak PRUNING & 27 & 0.044220 & 0.926 & 0.977 & 0.071 & 0.982 \\
VERTEBRAL\_\allowbreak COMPRESSION\_\allowbreak FRACTURE & 1,128 & 0.406460 & 0.967 & 0.970 & 0.714 & 0.994 \\
VERTEBRAL\_\allowbreak HEMANGIOMA & 113 & 0.068953 & 0.938 & 0.938 & 0.107 & 0.985 \\
\end{longtable}
\endgroup

\end{document}